\documentclass[letterpaper,journal]{IEEEtran}
\usepackage{amsmath,amsfonts,amssymb}
\usepackage{algorithmic}
\usepackage{algorithm}
\usepackage{array}
\usepackage{tabularx}
\usepackage[caption=false,font=normalsize,labelfont=sf,textfont=sf]{subfig}
\usepackage{textcomp}
\usepackage{stfloats}
\usepackage{url}
\usepackage[acronym,nomain,nonumberlist]{glossaries}
\usepackage{verbatim}
\usepackage{graphicx}
\usepackage{cite}
\usepackage{hyperref}
\usepackage{booktabs}
\usepackage{tabularx}
\usepackage{multirow}
\usepackage{etoolbox}

\makeatletter
\def\@IEEEBIOskipN{1\baselineskip}
\patchcmd{\IEEEbiography}
  {\vskip \@IEEEBIOskipN plus 1fil minus 0\baselineskip}
  {\vskip \@IEEEBIOskipN}
  {}{}
\makeatother
\setacronymstyle{long-short}
\newacronym{fmcw}{FMCW}{frequency-modulated continuous-wave}
\newacronym{mimo}{MIMO}{multiple-input multiple-output}
\newacronym{sbr}{SBR}{shooting and bouncing rays}
\newacronym{rcs}{RCS}{radar cross section}
\newacronym{adc}{ADC}{analog-to-digital converter}
\newacronym{gpu}{GPU}{graphics processing unit}
\newacronym{tdm}{TDM}{time-division multiplexing}
\newacronym{fmi}{FMI}{Functional Mock-up Interface}
\newacronym{osi}{OSI}{Open Simulation Interface}
\newacronym{ra}{RA}{range--angle}
\newacronym{rd}{RD}{range--Doppler}
\newacronym{rad}{RAD}{range--azimuth--Doppler}
\newacronym{fft}{FFT}{fast Fourier transform}
\newacronym{dft}{DFT}{Discrete Fourier Transform}
\newacronym{cfar}{CFAR}{Constant False Alarm Rate}
\newacronym{gan}{GAN}{generative adversarial network}
\newacronym{psf}{PSF}{point-spread function}
\newacronym{cir}{CIR}{channel impulse response}
\newacronym{cfr}{CFR}{channel frequency response}
\newacronym{rt}{RT}{ray tracing}
\newacronym{go}{GO}{geometrical optics}
\newacronym{po}{PO}{physical optics}
\newacronym{utd}{UTD}{uniform theory of diffraction}
\newacronym{sil}{SIL}{software-in-the-loop}
\newacronym{hil}{HIL}{hardware-in-the-loop}
\newacronym{vil}{VIL}{vehicle-in-the-loop}
\newacronym{rmse}{RMSE}{root-mean-square error}
\newacronym{psnr}{PSNR}{peak signal-to-noise ratio}
\newacronym{if}{IF}{intermediate-frequency}
\newacronym{itu}{ITU}{International Telecommunication Union}
\newacronym{hpd}{HPD}{Hit-Point Doppler}
\newacronym{fov}{FOV}{Field-Of-View}

\begin{document}

\title{AuRaSim: A Ray-Tracing Based Automotive \glsentryshort{fmcw} Radar Simulator for Dynamic Scenarios}

\author{Zhaoze~Wang$^{1,3}$,~\IEEEmembership{Graduate Student Member,~IEEE},
Changxu~Zhang$^{1,3}$,~\IEEEmembership{Graduate Student Member,~IEEE},
Tai~Fei$^{2}$,~\IEEEmembership{Senior Member,~IEEE},
Christopher~Grimm$^{1}$,~\IEEEmembership{Member,~IEEE},
Yi~Jin$^{1}$,~\IEEEmembership{Member,~IEEE},
Claas~Tebruegge$^{1}$,~\IEEEmembership{Member,~IEEE},
and~Markus~Gardill$^{3}$,~\IEEEmembership{Senior Member,~IEEE},

\thanks{Zhaoze Wang, Changxu Zhang, Christopher Grimm, Yi Jin, and Claas Tebruegge are with HELLA GmbH \& Co. KGaA, 59552 Lippstadt, Germany (e-mail:
\{zhaoze.wang, changxu.zhang, christopher.grimm, yi.jin, claas.tebruegge\}@forvia.com).}%

\thanks{Tai Fei is with the Department of Information Technology, Dortmund
University of Applied Sciences and Arts, 44139 Dortmund, Germany (e-mail:
tai.fei@fh-dortmund.de).}%

\thanks{Markus Gardill is with the Department of Electronic Systems and Sensors,
Brandenburg University of Technology, 03046 Cottbus, Germany (e-mail:
markus.gardill@b-tu.de).}%

\thanks{Manuscript received 2026.}}

\markboth{AuRaSim: A Ray-Tracing Based Automotive \glsentryshort{fmcw} Radar Simulator for Dynamic Scenarios, 2026}{Wang \MakeLowercase{\textit{et al.}}: AuraSim}

\maketitle

\begin{abstract}
Modern simulators can generate synthetic camera and LiDAR data with accurate annotations for downstream tasks. In contrast, automotive radar simulation remains less mature because it requires not only a 3D engine to model dynamic objects in the scenes but also electromagnetic material properties, antenna configurations, wave propagation mechanisms, and coherent signal synthesis. Existing frameworks are either proprietary or overly simplified, limiting the scalability of radar raw measurements. This paper presents AuRaSim, a ray-tracing-based automotive \gls{fmcw} radar simulation framework for dynamic scenarios. Rather than implementing all components from scratch, AuRaSim builds on Blender for efficient dynamic-scene construction and Sionna RT for \gls{gpu}-accelerated electromagnetic ray tracing, while introducing radar-specific adaptations for path-resolved propagation with complex coefficients, motion-aware Doppler modeling, and subsequently coherent multi-channel \gls{adc} cube synthesis based on waveform configurations.  Quantitative comparison between simulated and measured range--Doppler maps in digital-twin scenarios achieves a \gls{rmse} of 0.147 and a \gls{psnr} of 16.93~{dB}, demonstrating the accuracy and fidelity of the synthetic signal. AuRaSim therefore provides a complete and extensible platform for reproducible automotive radar simulation and research. Project homepage: \url{https://wangzhaoze.github.io/aurasim}.
\end{abstract}


\glsresetall

\section{Introduction}

\IEEEPARstart{A}{utomotive} \gls{fmcw} radar is a crucial sensing modality for autonomous driving because it provides direct velocity measurements and remains robust under adverse environmental conditions, where camera- and
LiDAR-based perception may degrade~\cite{kong2024survey}. However, radar measurements are typically sparse, noisy, affected by the multipath effect, and have limited angular resolution, which challenges conventional signal processing algorithms in complex traffic situations. 

Recent deep learning based methods have shown promising performance in overcoming some of these limitations, but their development is still constrained by the lack of large-scale and accurately annotated radar datasets. Compared with camera and LiDAR data, radar raw signals are more challenging to interpret and annotate directly because they require specialized domain knowledge, making high-quality ground-truth acquisition much more expensive. Although several automotive radar datasets have been released in recent years (e.g., \cite{raddet,kradar}), their scale and annotation variety remain more limited than widely used camera- and LiDAR-based benchmarks such as KITTI \cite{kitti} and nuScenes \cite{nuscenes}.

Existing approaches to radar data synthesis can broadly be summarized as data-driven and physics-based methods. Data-driven approaches learn the distribution of recorded data and synthesize radar representations using conditional generative models \cite{de_oliveira_generating_2020,wang_synthetic_fmcw_2026}. Such methods can efficiently reproduce measurement statistics and sensor effects that are difficult to formulate analytically. Their applicability is nevertheless often hardware-specific, because the learned distribution is inherently conditioned on the waveform and antenna layout represented during training, such that deviations from these settings can induce domain shifts and degrade generalization to unseen radar hardware. Physics-based simulation instead derives radar measurements from explicit scene, hardware, and propagation models \cite{schasler_realistic_2021}. This enables controlled configuration variation while providing exact scene state as ground truth, making it attractive for radar dataset generation.

However, building a complete automotive radar simulator remains challenging for several reasons. First, realistic driving scenarios require a dynamic 3D scene representation, including environmental geometry, vehicle movements, and articulated or deformable road users. Second, geometrically accurate 3D assets must be assigned electromagnetic material properties for roads, vehicles, buildings, and other traffic participants. Third, relevant electromagnetic propagation mechanisms, including specular and diffuse reflection, diffraction, and multipath propagation, must be modeled to reproduce realistic radar returns, including distributed environmental echoes and multipath-induced ghost targets. Fourth, noise, calibration errors, interference, and other sensor effects remain difficult to reproduce accurately. Finally, resolving these interactions over multiple antenna channels and successive radar frames is computationally expensive. Consequently, automotive radar simulation requires careful trade-offs among physical fidelity, computational efficiency, and practical scenario construction.

Current simulation tools address these challenges to different degrees, with many approaches relying on simplifying assumptions. General-purpose autonomous-driving simulators such as CARLA
\cite{dosovitskiy_carla_2017} provide dynamic traffic environments and radar
detections but use simplified ray-casting models. Similarly, MATLAB employs a
measurement-level probabilistic radar model to generate detections
\cite{mathworks_radar_generator}. Commercial radar simulation tools such as
Remcom WaveFarer \cite{wavefarer}, Altair Feko \cite{altair_feko} and dSPACE AURELION
\cite{dspace_aurelion} provide higher-fidelity radar simulation. These platforms demonstrate that physically detailed automotive radar simulation is feasible, but their proprietary and closed-source nature limits wide application within the academic community. This motivates an accessible and extensible framework for flexible research and efficient synthetic-data generation.

In this work, we present AuRaSim, an end-to-end automotive radar simulation framework that addresses this gap by building upon and substantially extending the capabilities of two complementary open-source technologies rather than implementing a complete rendering and propagation stack from scratch. Blender \cite{blender} provides a mature solution for constructing complex polygonal scenes, managing reusable 3D assets, i.e., object models and materials, and representing time-varying animations. Sionna RT is an open-source ray tracer originally developed for radio-propagation modeling in communication systems and provides electromagnetic material models and path solvers supporting specular and diffuse reflection, diffraction, and transmission \cite{sionna}. AuRaSim integrates the complementary capabilities of Blender~\cite{blender} and Sionna RT~\cite{sionna} and substantially extends them to form an end-to-end automotive FMCW radar simulation framework.
 
Our main contributions are threefold: (1) We develop a unified scenario-to-signal simulation pipeline that produces path-resolved propagation data, multi-channel \gls{adc} samples, and derived radar products while preserving a consistent representation of scene state, sensor geometry, and waveform timing. (2) We integrate radar-specific adaptations to the underlying propagation pipeline, including field-of-view-aware ray sampling, propagation-path filtering, and a vertex velocity field that enables \gls{hpd} modeling for complex dynamic objects. The effectiveness of these components is evaluated through dedicated ablation studies. (3) We establish a digital-twin-based validation workflow by reconstructing representative real-world measurement scenarios with matched 77-GHz radar configurations and quantitatively comparing simulated and measured radar data, including range--Doppler representations, to evaluate the signal fidelity of the proposed simulator.

\section{Related Work}

\subsection{Physics-Based Radar Simulation}
\label{sec:related_physics}

Physics-based radar simulation derives sensor measurements from explicit models of scene geometry, electromagnetic propagation, and radar signal formation. Hirsenkorn \emph{et al.} \cite{hirsenkorn_ray_2017} presented an automotive \gls{fmcw} radar simulator based on geometrical-optics ray launching, including antenna patterns, reflections, and ray divergence on curved surfaces, comparing range profiles between simulated and measured data. High-fidelity physics-based automotive radar simulation commonly employs the \gls{sbr} method introduced in \cite{ling_sbr_1989}. Sch\"{u}ssler \emph{et al.} \cite{schasler_realistic_2021} extended \gls{sbr}-based simulation to large \gls{mimo} arrays and evaluated simulated range-angle spectrum on single-chirp measurements in static scenes and therefore primarily investigated spatial radar responses without Doppler. Mergenthaler \emph{et al.} \cite{mergenthaler_microdoppler_2025} later extended this line of work to simulate dynamic pedestrians by incorporating motion-captured animation in a controlled indoor scenario. Thieling \emph{et al.} \cite{thieling_scalable_2021} proposed a modular architecture that includes physical interactions, scene geometry, object motion, and atmospheric attenuation by rain, with particular emphasis on computational scalability and real-time execution, while detailed implementations of several submodels and quantitative sim-to-real validation are not reported. ViRa \cite{schoffmann_virtual_2021} follows a different approach by using \gls{gpu}-based rendering and multi-depth buffers rather than a conventional \gls{sbr} solver to generate \gls{fmcw} radar raw data. It supports multi-antenna configurations, configurable beam patterns, and penetration through non-conductive objects, and demonstrates good agreement with measurements. However, its evaluation focuses primarily on indoor robotics scenarios and does not address dynamic automotive traffic scenes or detailed motion-induced Doppler effects. More recently, RadaRays \cite{mock_radarays_2025} introduced an open-source ray-tracing framework for rotating \gls{fmcw} radars in mobile robotics. It models reflection, transmission, and material-dependent scattering, outputs range-azimuth responses and provides a substantially more physical alternative to the LiDAR-like ray-casting radar models. However, its output remains at the range-azimuth level rather than providing coherent multi-channel ADC signals, which are required for waveform-level \gls{mimo} and Doppler processing. Haider \emph{et al.} couple CarMaker ray-tracing outputs with an automotive \gls{fmcw} baseband model including virtual \gls{mimo} and sensor impairments \cite{haider_modeling_2025}. Compared to those existing approaches, AuRaSim provides a unified scenario-to-signal pipeline for dynamic scenes, physics-based propagation, and coherent multi-channel \gls{adc} synthesis.

\subsection{Generative Radar Synthesis}
Data-driven methods typically learn the distribution of a selected radar measurement representation. 
L2R \cite{wang_l2r_2020} translates a LiDAR bird's-eye-view representation into a radar spectrum image using a \gls{gan},  although phase and channel information absent from the LiDAR input cannot be recovered.
Fidelis \emph{et al.} \cite{fidelis_generation_2023} operate at a lower signal level by generating 16 raw chirps conditioned on motorcycle distance and Gaussian noise, while their evaluation is limited to a single motorcycle moving along a straight trajectory. 
RF Genesis \cite{chen_rf_2023} maps visual scene representations to corresponding mmWave sensing data by using a set of diffusion models to generate large numbers of 3D scenes from application-specific prompts. 
RadarGen \cite{borreda_radargen_2026}, in contrast, generates processed automotive radar point clouds from multi-view imagery together with visual depth, semantic, and motion priors, thereby inheriting errors from the visual cues while not preserving low-level phase or channel information. 
RadSimReal \cite{bialer_radsimreal_2024} converts material-assigned reflection points into an intensity radar tensor using a measured \gls{psf} and noise model,  but does not generate time-domain ADC data. 
Xiao \emph{et al.} \cite{xiao_simulate_2025} learn an attribute-conditioned \gls{rad} tensor from a reflection tensor and waveform-parameter embedding. The resulting output is a real-valued \gls{rad} cube rather than a time-domain \gls{adc} cube or complex \gls{mimo} channel tensor, limiting access to channel-dependent phase information. 
De Oliveira and Bekooij \cite{de_oliveira_generating_2020} synthesize short-range \gls{fmcw} \gls{rd} maps using a bounding-box-conditioned \gls{gan}. 
Wang \emph{et al.} \cite{wang_synthetic_fmcw_2026} propose a conditional diffusion model for automotive range--azimuth map generation and subsequently extend it to temporally consistent sequences using a Video Diffusion Transformer \cite{wang_temporal_vdt_2026}. However, these methods learn radar data distributions specific to the sensor configurations represented in datasets, limiting their generalization to customized antenna arrays and waveform parameters. In contrast, AuRaSim supports configurable radar hardware and generates coherent multi-channel ADC signals through physics-based simulation.

\subsection{Radar Neural Fields}

Neural radar scene representations provide an alternative approach to radar data synthesis by reconstructing a previously observed environment from radar measurements and rendering radar observations from novel sensor views. Radar Fields models frequency-space \gls{fmcw} measurements using occupancy and view-dependent reflectance fields, but assumes static targets and zero Doppler \cite{borts_radar_2024}. DART \cite{huang_dart_2024} incorporates range--Doppler measurements and exploits sensor motion to reduce angular ambiguity, but remains limited to static scenes and is sensitive to velocity-estimation error. RF4D \cite{RF4D} represents an outdoor dynamic scene with time-conditioned occupancy and \gls{rcs} fields, but its input and output are 2-D range--azimuth scans rather than \gls{adc} data. RadarSplat \cite{kung_radarsplat_2025} adds Gaussian primitives and models saturation, speckle, and multipath in scanning-radar images. Its current formulation is explicitly restricted to static scenes. RadarSim \cite{chen_radarsim_2026} uses synchronized RGB and single-chip radar recordings to render \gls{rd} frames from a camera-initialized neural field. These methods synthesize novel views of reconstructed scenes, whereas AuRaSim performs path-resolved simulation of newly constructed dynamic driving scenarios.

\subsection{Automotive Radar Simulation Software}

Radar simulation tools span a broad range of abstraction levels, from simplified functional models that directly generate detections to physically based simulators that model electromagnetic propagation and raw ADC signals. CARLA \cite{dosovitskiy_carla_2017} employs conical ray casting and directly returns detections characterized by range, angle, and relative velocity, while MATLAB \cite{mathworks_radar_generator} uses statistical measurement-level models based on target signatures and detection probability. Commercial tools provide further high-fidelity capabilities. Ansys HFSS \glsentryshort{sbr}+ \cite{ansys_hfss_sbr} supports \gls{sbr}-based electromagnetic analysis and automotive range--Doppler simulation. Remcom WaveFarer \cite{wavefarer} combines ray-based propagation with \gls{po} and diffraction models and produces outputs including \gls{cir} and I/Q data. dSPACE AURELION \cite{dspace_aurelion} provides radar-optimized materials, multipath ray tracing, and outputs ranging from \gls{cir} and raw radar data to detection and object lists for real-time X-in-the-loop applications. CarMaker Radar RSI \cite{ipg_carmaker_rsi} integrates physically based radar-environment modeling into dynamic driving scenarios. RadarSimPy \cite{radarsimpy} combines configurable radar waveforms with point and mesh targets to generate complex baseband samples through ray tracing, while its \gls{gpu}-accelerated backend is distributed as a precompiled licensed implementation. These tools provide high-fidelity simulation capabilities, but proprietary or licensed implementations can restrict inspection and modification of the underlying propagation and signal-generation algorithms. AuRaSim instead provides an accessible and extensible research framework for automotive \gls{fmcw} radar studies.

\section{Preliminaries}
\label{sec:preliminaries}

\begin{figure*}[t]
    \centering
    \includegraphics[width=0.99\linewidth]{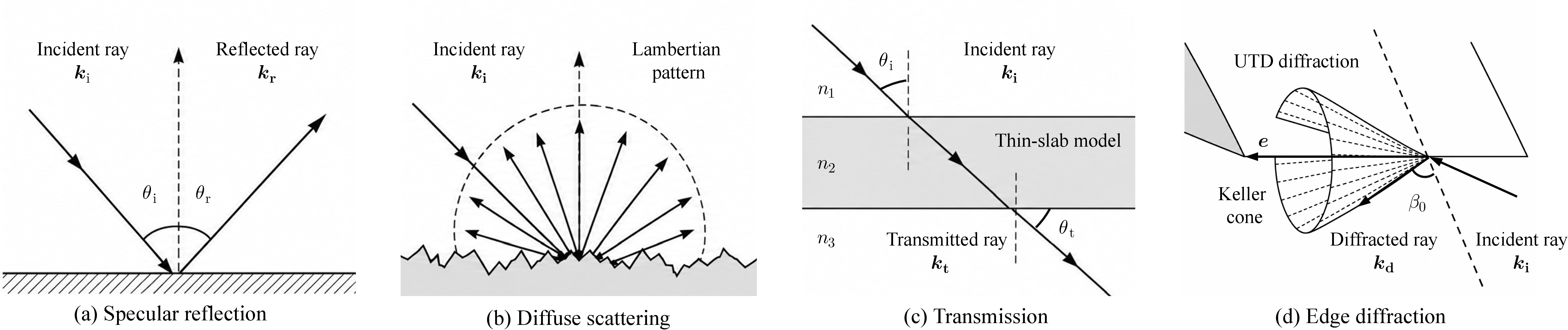}
    \caption{Principal propagation mechanisms considered by the Path Solver:
    specular reflection, diffuse scattering, thin-slab
    transmission, and edge diffraction.}
    \label{fig:em_mechanisms}
\end{figure*}

\subsection{FMCW Radar Signal Processing}
\label{sec:fmcw_prelim}

An \gls{fmcw} radar estimates target range and relative radial velocity from the
frequency and phase evolution of the dechirped received signal. The complex
transmit signal under the  linear \gls{fmcw} signal model~\cite{stove_linear_1992} can be written as
\[\
    s_{\mathrm{tx}}(t)
    =
    \exp
    \left\{
        \jmath 2\pi
        \left(
            f_ct+\frac{\mu}{2}t^2
        \right)
    \right\},
    \qquad 0\leq t<T_c ,
    \label{eq:prelim_tx_fmcw}
\]
where $s_{\mathrm{tx}}(t)$ denotes the complex transmitted waveform, $t$ is
the fast-time variable within one chirp, $f_c$ is the carrier frequency,
$\mu$ is the linear frequency slope, and $T_c$ denotes the chirp duration.
The imaginary unit is denoted by $\jmath$.

For a point target at range $R$ moving with relative radial velocity $v$, the round-trip propagation delay $\tau$ and Doppler frequency $\nu$ over one chirp are approximately
\[
    \tau=\frac{2R}{c_0},
    \qquad
    \nu=\frac{2v}{\lambda_0},
    \label{eq:prelim_delay_doppler}
\]
where $c_0$ is the speed of light, and
$\lambda_0=c_0/f_c$ is the wavelength at the carrier frequency. Accordingly, the received signal can be modeled as
\[
    s_{\mathrm{rx}}(t)
    =
    A\,
    s_{\mathrm{tx}}(t-\tau)
    e^{\jmath 2\pi\nu t},
\]
where $A$ denotes the complex path coefficient accounting for propagation loss, antenna responses, and electromagnetic interactions such as reflection and scattering. After dechirping and neglecting the quadratic delay term $-\mu\tau^2/2$, which is negligible for short-range automotive radar when $\tau/T_c \ll 1$ \cite{xu_difference_2021}, the corresponding \gls{if} signal is modeled as
\begin{equation}
    s_{\mathrm{IF}}(t)
    =
    A
    \exp
    \left\{
        \jmath 2\pi
        \left[
            \left(
                \mu\tau-\nu
            \right)t
            +
            f_c\tau
        \right]
    \right\},
    \label{eq:prelim_if_signal}
\end{equation}
where the term $\mu\tau-\nu$ determines the beat frequency, whereas
$f_c\tau$ represents the delay-dependent carrier phase. 

After analog-to-digital conversion through \gls{adc}, the \gls{if} signal is sampled along the
fast-time dimension within each chirp, while successive chirps form the
slow-time dimension. The phase evolution in the fast-time signal mainly contains range-dependent
beat-frequency information, whereas the phase progression across slow time
encodes Doppler frequency. For multi-channel radar systems, the relative phase between the complex-valued \gls{if} signals across antenna channels additionally contains angular
information. These dimensions jointly form the multi-channel ADC cube, which is formed by coherently summing the beat-signal contributions of all resolved propagation paths while preserving their relative phases across samples, chirps, and antenna channels. The corresponding path-to-ADC synthesis and the temporal approximation used
in AuRaSim are described in detail in
Sec.~\ref{sec:adc_synthesizer}.

\subsection{Sionna RT Path Solver}
\label{sec:sionna_prelim}

This subsection summarizes the Sionna RT propagation formulation relevant to
AuRaSim, while further implementation details are provided in \cite{sionna}.

\subsubsection{Ray Sampling}
\label{sec:ray_sampling}

Spherical Fibonacci lattice sampling has been adopted in previous works \cite{thieling_scalable_2021, hirsenkorn_ray_2017} to initialize \gls{sbr} rays. Compared with a uniform
angular grid, directly sampling azimuth and polar angles with constant angular increments leads to nonuniform point density on the sphere, with excessive
samples near the poles. Fibonacci sampling instead provides a deterministic
and approximately equal-area distribution over the sphere with reduced clustering, allowing
a given number of rays to cover the angular domain more efficiently.

For $N_{\mathrm{ray}}$ rays, the spherical Fibonacci angles are
defined as

\begin{equation}
    \theta_n
    =
    \arccos\left(
        1-\frac{2n}{N_{\mathrm{ray}}-1}
    \right),
    \phi_n
    =
    2\pi\,
    \mathrm{frac}\left(
        \frac{n}{\phi_g}
    \right)
    \label{eq:fibonacci_theta}
\end{equation}
where $n = 0, \ldots, N_{\mathrm{ray}} - 1$, $\phi_g = (1+\sqrt{5})/2$
denotes the golden ratio, and
$\operatorname{frac}(x) = x - \lfloor x \rfloor$
denotes the fractional part of $x$. The corresponding initial unit propagation direction of the $n$-th ray is

\begin{equation}
    {\boldsymbol{k}}_n^{(0)}
    =
    \begin{bmatrix}
        \sin\theta_n\cos\phi_n\\
        \sin\theta_n\sin\phi_n\\
        \cos\theta_n
    \end{bmatrix}.
    \label{eq:sionna_fibonacci}
\end{equation}

This distribution covers the complete sphere, such that each initial ray tube
represents an approximately equal solid angle

\[
    \Delta\Omega_{\mathrm{full}}
    \approx
    \frac{4\pi}{N_{\mathrm{ray}}}.
\]

\subsubsection{SBR Candidate Generation}
\label{sec:sbr_generation}

\gls{sbr} traces rays launched from the transmitter through successive interactions with the scene geometry to identify candidate propagation paths. To avoid the exponential growth that would result from spawning all physically possible continuations at every intersection, at most one interaction type is sampled for each ray and path depth. At the $\ell$-th propagation interaction, the selected type $\chi^{(\ell)}$ is drawn from the set $\mathcal{X}$ with probability $p_\chi$ as
\[
    \mathcal{X}
    =
    \{\mathcal{R},\mathcal{S},\mathcal{T},\mathcal{D}\},
    \quad \text{s.t.} 
    \quad \sum_{{\chi^{(\ell)}}\in\mathcal{X}} p_{\chi^{(\ell)}}=1, \quad p_{\chi^{(\ell)}}\geq0, 
    \label{eq:interaction_sampling}
\] 
where specular reflection $\mathcal{R}$, diffuse scattering $\mathcal{S}$, and transmission $\mathcal{T}$ are importance-sampled from the material response and Fresnel coefficients, while diffraction $\mathcal{D}$ is assigned a separately specified sampling probability to ensure sufficient exploration of weak diffracted paths.

\subsubsection{Electromagnetic Propagation Mechanisms}
\label{sec:em_interactions}

The four interaction mechanisms considered by the Path Solver are illustrated
in Fig.~\ref{fig:em_mechanisms}. They affect both the geometry of the
propagation path and its complex electromagnetic field.

\paragraph{Specular Reflection}

For a smooth surface with unit normal $\boldsymbol{n}$, the outgoing ray
direction follows the geometrical-optics reflection law
\[
    {\boldsymbol{k}}_{\mathrm r}
    =
    {\boldsymbol{k}}_{\mathrm i}
    -
    2
    \left(
        {\boldsymbol{k}}_{\mathrm i}^{\mathsf T}\boldsymbol{n}
    \right)
    \boldsymbol{n},
\]
where ${\boldsymbol{k}}_{\mathrm{i}}$ and
${\boldsymbol{k}}_{\mathrm{r}}$ denote the incident and reflected
unit propagation directions. The reflected field is obtained by applying the polarization-dependent complex Fresnel response of the assigned radio material to the incident field. 

\paragraph{Diffuse Scattering}

Sionna RT models diffuse reflection using an effective-roughness formulation, where the scattering coefficient $S\in[0,1]$ determines the fraction $S^2$ of the total reflected power assigned to diffuse scattering, while the normalized scattering pattern $f_{\mathrm{s}}$ specifies its angular distribution over the outward hemisphere $\mathcal{H}^{+}$,
\[
    \int_ {\mathcal{H}^{+}}
        f_{\mathrm{s}}
        \left(
            {\boldsymbol{k}}_{\mathrm{i}},
            {\boldsymbol{k}}_{\mathrm{s}}
        \right)
        \mathrm{d}\Omega
    =1.
    \label{eq:diffuse_normalization}
\]

Sionna RT provides several scattering patterns based on effective-roughness models \cite{degli_esposti_scattering_2007}. The Lambertian model is used by default and distributes the scattered energy around the surface normal according to
\[
    f_{\mathrm{L}}(\theta_{\mathrm{s}})
    =
    \frac{\cos\theta_{\mathrm{s}}}{\pi},
    \qquad
    \theta_{\mathrm{s}}\in[0,\pi/2], 
    \label{eq:lambertian_pattern}
\]
where $\theta_{\mathrm{s}}$ denotes the angle between the scattering direction and the surface normal.
The directive model instead concentrates the scattered field around the ideal specular-reflection direction, with its lobe width controlled by a directivity parameter. The backscattering model further introduces an additional lobe toward the incident direction, enabling enhanced return toward the source.

Unlike specular reflection, diffuse scattering distributes energy over a continuum of outgoing directions. Sionna RT handles this using next-event estimation, which explicitly tests the connection from each diffuse interaction point to the receiver. If the receiver is visible, the corresponding receiver-directed contribution is retained as a candidate path and weighted according to the scattering model. An outgoing direction is then sampled stochastically from the scattering distribution to continue the \gls{sbr} trajectory and explore higher-order propagation paths, while the corresponding interaction-sampling and ray-tube solid-angle weights are applied during field evaluation to compensate for the stochastic sampling and properly normalize the resulting path contributions.

\paragraph{Transmission}

Sionna RT models transmissive surfaces using a homogeneous thin-slab approximation. The effective slab transmission coefficient is expressed as

\[
    T_q(\delta)
    =
    \frac{
        (1-r_q^2)e^{-{\jmath} \psi}
    }{
        1-r_q^2e^{-\jmath2\psi}
    },
    \qquad
    \psi
    =
    k_0\delta
    \sqrt{
        \widetilde{\varepsilon}_r-\sin^2\theta_{\mathrm i}
    } .
\]

Here, $q$ indexes the two polarization components with respect to the plane of incidence, i.e., perpendicular ($\perp$) and parallel ($\parallel$), and $r_q$ denotes the corresponding single-interface Fresnel reflection coefficient. Moreover, $\delta$ is the effective material thickness, $k_0=2\pi/\lambda_0$ is the free-space wavenumber, and $\widetilde{\varepsilon}_r$ is the complex relative permittivity determined by the material permittivity and conductivity. The numerator accounts for transmission through the two slab interfaces and propagation through the slab, whereas the denominator captures the effect of internal multiple reflections.

\paragraph{Edge Diffraction}

Geometrical optics alone cannot represent fields propagating around sharp
edges into geometrical shadow regions. Sionna RT therefore models edge
diffraction using the \gls{utd}~\cite{kouyoumjian_uniform_1974}. For an edge with unit direction ${\boldsymbol{e}}$, the incident and diffracted rays satisfy the Keller-cone condition
\[
    {\boldsymbol{k}}_{\mathrm i}^{\mathsf T}{\boldsymbol{e}}
    =
    {\boldsymbol{k}}_{\mathrm d}^{\mathsf T}{\boldsymbol{e}}
    =
    \cos\beta_0 .
\]
where $\beta_0$ denotes the angle between the incident ray and the
diffracting edge. The \gls{utd} diffraction coefficient determines the
complex amplitude and phase of the diffracted field as a function of
wavelength, wedge geometry, incidence and observation directions, and the
electromagnetic properties of the wedge faces. \gls{utd} transition functions
provide a continuous field description across geometrical shadow
boundaries, and only first-order diffraction
is considered in Sionna RT.

\subsubsection{Image-Method Path Refinement}
\label{sec:image_refinement}

The \gls{sbr} stage primarily identifies candidate interaction sequences rather than exact propagation paths between transmitter and receiver. While diffuse interactions can directly connect to a visible receiver through next-event estimation, randomly sampled specular, transmitted, or diffracted rays have essentially zero probability of intersecting a point receiver. Sionna RT therefore uses an image-method-based refinement to determine the exact interaction points of suitable candidate sequences such that they connect the transmitter and receiver and satisfy the corresponding propagation constraints.

For specular chains, the source is successively mirrored across the candidate reflection planes, so that the multi-bounce path can be solved as an equivalent straight-line connection to the receiver. The reflection points are subsequently obtained by tracing this connection back through the corresponding surfaces. Transmission events are included without angular deflection under the thin-slab approximation, while candidates containing first-order diffraction are handled in image space by additionally constraining the diffraction point to the corresponding wedge. Candidates whose refined interaction points lie outside the associated primitives or whose paths are occluded are discarded.
 
\subsubsection{Electromagnetic Field Calculation}
\label{sec:field_calculation}

After the geometric path has been validated, Sionna RT evaluates the complex
electromagnetic field along the complete interaction sequence. Starting from
the transmit antenna response, every interaction applies a complex
polarization-dependent transfer matrix
$\mathbf{T}_{p,\ell}$ corresponding to reflection, diffuse scattering,
transmission, or diffraction. The receive antenna pattern is subsequently
evaluated in the final arrival direction.

The complex path coefficient can be summarized as
\[
    \resizebox{0.98\columnwidth}{!}{$\displaystyle
    A_p
    =
    \underbrace{
        \frac{\lambda_0}{4\pi}
        G_p
    }_{\substack{\text{geometric spreading}\\\text{and normalization}}}
    \;
    \underbrace{
        \boldsymbol{F}_{\mathrm{rx}}^{\mathrm H}
        \left(
            -{\boldsymbol{k}}_p^{(L)}
        \right)
    }_{\text{receive antenna}}
    \;
    \underbrace{
        \left[
            \prod_{\ell=1}^{L}
            \mathbf{T}_{p,\ell}
        \right]
    }_{\substack{\text{material and}\\\text{EM interactions}}}
    \;
    \underbrace{
        \boldsymbol{F}_{\mathrm{tx}}
        \left(
            {\boldsymbol{k}}_p^{(0)}
        \right)
    }_{\text{transmit antenna}},
    $}
\]
where $\boldsymbol{F}_{\mathrm{tx}}$ and $\boldsymbol{F}_{\mathrm{rx}}$ denote the
complex transmit and receive antenna field responses, and $(\cdot)^{\mathrm H}$
denotes the Hermitian transpose. The interaction matrices
$\mathbf{T}_{p,\ell}$ are applied in propagation order from the transmitter
toward the receiver. The factor $G_p$ denotes the path-dependent geometric spreading factor, including the normalization required for stochastic ray sampling. For diffuse paths, $G_p$ additionally accounts for the interaction-sampling and ray-tube weights. For diffracted paths, the corresponding \gls{utd} spreading factor is used.

\section{Radar Simulation System}
\label{sec:radar_simulation_system}

AuRaSim provides a complete automotive radar simulation chain for dynamic scenarios. Following the system overview in Sec.~\ref{sec:system_overview}, antenna characteristics and waveform parameters are defined in Sec.~\ref{sec:sensor_config}. The Blender Scenario Builder (Sec.~\ref{sec:scenario_builder}) constructs and updates the dynamic scene. The Path Solver (Sec.~\ref{sec:path_tracer}) resolves material- and motion-dependent propagation paths. Finally, the \gls{adc} Signal Synthesizer (Sec.~\ref{sec:adc_synthesizer}) converts these paths into radar signals. This modular design preserves traceability throughout the simulation pipeline while allowing individual stages to be inspected and reused independently.

\subsection{System Overview}
\label{sec:system_overview}

\begin{figure*}[t]
    \centering
    \includegraphics[width=\textwidth]{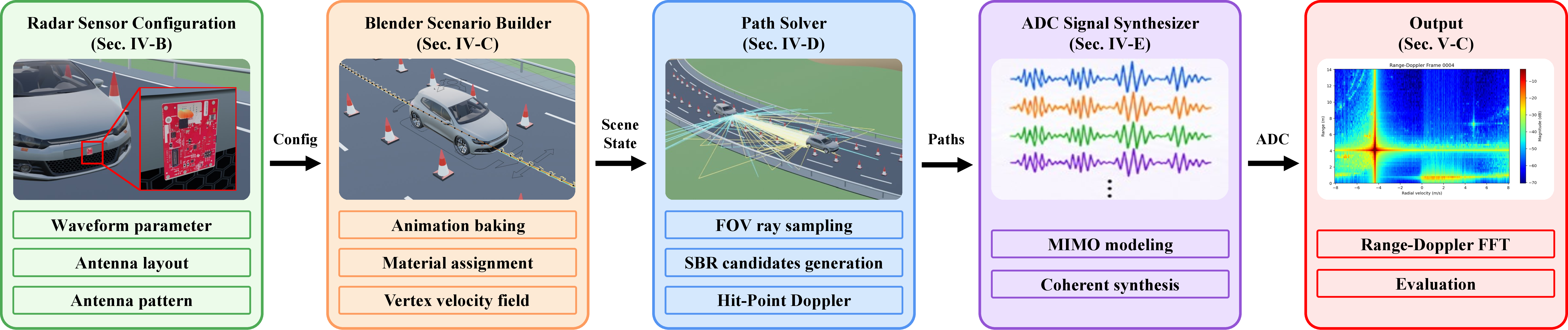}
    \caption{System overview of AuRaSim. A trajectory-driven dynamic
    digital twin and a configurable radar sensor are synchronized at each
    radar timestamp. The Path Solver produces an explicit path-level
    representation, which is subsequently converted into waveform-consistent
    multi-channel ADC samples and derived radar products.}
    \label{fig:system_overview}
\end{figure*}

As illustrated in Fig.~\ref{fig:system_overview}, AuRaSim maps a dynamic
digital-twin scene and a configurable radar sensor to multi-channel radar ADC
signals. The Radar Sensor Configuration defines the antenna layout,
radiation patterns, waveform, and acquisition parameters, while the
Blender Scenario Builder provides the time-varying scene geometry and
motion state. For each radar frame, the Path Solver resolves the path
set
$\mathcal{P}=\{P_{p}\}$, containing the complex coefficients, delays,
Doppler frequencies, and propagation geometry required by the
\gls{adc} Signal Synthesizer, which subsequently converts
these paths into a complex multi-channel ADC cube according to the configured
waveform and sampling scheme. Retaining $\mathcal{P}$ decouples the propagation stage from waveform-dependent ADC synthesis.


\subsection{Radar Sensor Configuration}
\label{sec:sensor_config}

AuRaSim provides a parameter-configurable radar sensor model that consists of three main components: \gls{mimo} setup, antenna radiation patterns, and waveform parameters. 

\subsubsection{MIMO Configuration}

The antenna layout specifies the number and relative positions of the
transmitters and receivers by
$\boldsymbol{d}^{\mathrm{tx}}_u\in\mathbb{R}^3$ and
$\boldsymbol{d}^{\mathrm{rx}}_r\in\mathbb{R}^3$, respectively. AuRaSim supports
both explicit-array tracing, in which propagation is evaluated for individual
transmit--receive positions, and a reference-array approximation, in which propagation is traced between the array reference positions and the responses of individual transmit--receive channels are reconstructed using element-dependent phase offsets derived from the departure and arrival directions. For \gls{tdm}-\gls{mimo}, the transmit sequence is additionally retained to preserve
the temporal relationship between virtual-array channels. The corresponding
channel synthesis is described in Sec.~\ref{sec:adc_synthesizer}.

\subsubsection{Antenna Pattern}

The antenna radiation pattern is represented by the direction-dependent complex field response $\boldsymbol{F}(\theta,\phi)$, where $\theta$ and $\phi$ denote the polar and azimuth directions. Customized antenna patterns can be interpolated at the propagation directions required by the ray tracer. The resulting antenna responses are incorporated into the electromagnetic field calculation described in Sec.~\ref{sec:field_calculation}.

\subsubsection{Waveform Parameters}

Signals with different waveforms can be synthesized by specifying the
corresponding transmit waveform and sampling scheme. In this work, we focus
on automotive \gls{fmcw} radar. The configurable waveform parameters include
the carrier frequency $f_c$, effective bandwidth $B$, chirp slope $\mu$, chirp
and idle durations, ADC sampling frequency $f_s$, number of ADC samples per
chirp $N_s$, and number of chirps per radar frame $N_c$. 

\subsection{Blender Scenario Builder}
\label{sec:scenario_builder}

The Blender Scenario Builder constructs plausible dynamic driving scenes from trajectory-level descriptions and provides the downstream Path Solver with synchronized geometry and mesh motion state. Given object trajectories and semantic classes, corresponding 3D assets are retrieved from the asset library, placed into the scene, and animated according to their motion over time. We use Blender\cite{blender} as the digital-twin front end because it provides mature support for polygon meshes, asset management, rigid-body motion, skeletal animation, and evaluated deformable geometry.

\subsubsection{Dynamic Scene Representation and Persistent Update}

AuRaSim partitions the scene geometry into three categories,
\[
    \mathcal{G}(t)
    =
    \mathcal{G}_{\mathrm{stat}}
    \cup
    \mathcal{G}_{\mathrm{rig}}(t)
    \cup
    \mathcal{G}_{\mathrm{def}}(t),
    \label{eq:geometry_classes}
\]
where $\mathcal{G}_{\mathrm{stat}}$ denotes static infrastructure,
$\mathcal{G}_{\mathrm{rig}}(t)$ rigid dynamic objects, and
$\mathcal{G}_{\mathrm{def}}(t)$ articulated or deformable objects.
Typical examples include roads and buildings, vehicles, and pedestrians,
respectively.

Blender evaluates the complete animation state at each
requested radar time $t_k$. Scene semantics are used to select the corresponding mesh and
animation configuration from the 3D asset library. Vehicle assets are animated using a Blender vehicle-rigging workflow such as Rigacar \cite{rigacar}, allowing the vehicle body, steering, and wheel
rotation to be synchronized with the trajectory. Articulated human meshes
use skeletal animations, e.g., motion assets from Adobe Mixamo
\cite{adobe_mixamo}, adapted to
the specified pedestrian trajectories. In this way, different parts of a dynamic
object are not restricted to a single object-level velocity, allowing more
plausible vehicle and pedestrian motion to be represented.

Repeatedly rebuilding and transferring the complete electromagnetic scene to \gls{gpu} for every radar frame would introduce substantial computational overhead.
To avoid this, AuRaSim introduces a persistent propagation-scene update strategy that treats static, rigid, and deformable geometry differently. Static geometry is instantiated once with simulation initialization,
rigid objects are updated through their transformations, and deformable
objects retain their triangle connectivity while only their evaluated vertex
positions are updated.

\subsubsection{Radio Material Assignment}

The 3D asset library stores electromagnetic material information independently
of the visual materials used for rendering. During scene construction, each
mesh is assigned a radio material according to its semantic and physical
properties. AuRaSim uses the radio-material representation provided by
Sionna RT \cite{sionna}, including its built-in \gls{itu} material
models for common materials such as concrete, glass, wood, brick, metal and
ground surfaces.

Each radio material specifies its electromagnetic properties, including
permittivity, conductivity, material thickness, diffuse-scattering coefficient,
cross-polarization coefficient, and scattering pattern. For the built-in \gls{itu} material models, the electromagnetic
properties are frequency dependent and are evaluated at the configured radar
frequency. The Path Solver subsequently uses these material properties
for material-dependent propagation interactions.

\begin{figure*}[t]
    \centering
    \includegraphics[width=0.98\textwidth]{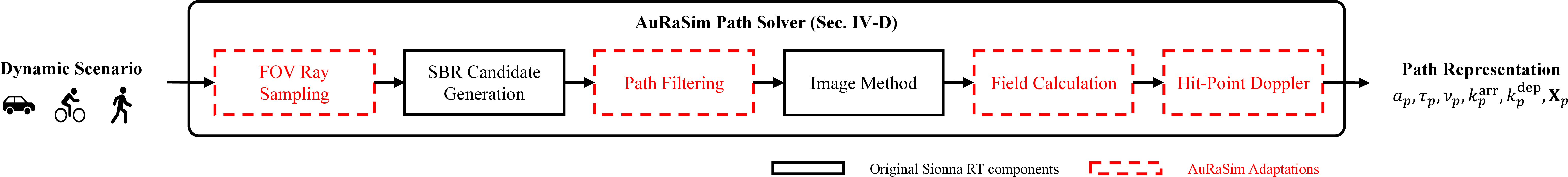}
    \caption{The Path Solver pipeline of AuRaSim. Black components correspond to the
    Sionna RT $\mathtt{PathSolver}$, while red components indicate the radar-oriented
    adaptations introduced in AuRaSim. AuRaSim restricts the initial ray
    distribution to the radar field of view, applies path filtering before
    field evaluation, provides local hit-point velocities for Doppler
    calculation, and exports an explicit path-level representation for
    subsequent radar signal synthesis.}
    \label{fig:path_tracer}
\end{figure*}

\subsubsection{Vertex-Velocity Field}
\label{sec:vertex_velocity}

Sionna RT represents the motion of each dynamic scene object by a single
velocity vector $\boldsymbol{v}(o)$, which is shared by all propagation
interactions occurring on that object \cite{sionna}.
While this representation is suitable for objects undergoing approximately
uniform translational motion, it cannot capture spatially varying surface
velocities caused by rotation, articulation or deformation.

For automotive radar simulation, such local motion is particularly relevant.
Points on a rotating wheel exhibit different instantaneous velocities from
the car body, while the limbs of an articulated pedestrian undergo
additional motion relative to the pedestrian trajectory. To capture these effects, AuRaSim introduces a world-space vertex-velocity field for the dynamic scene geometry that can be queried at individual interactions. 

For rigid objects, the velocity of vertex $i$ is obtained from the
translational and rotational motion of the object,
\begin{equation}
    \boldsymbol{v}_i(t_k)
    =
    \boldsymbol{v}_o(t_k)
    +
    \boldsymbol{\omega}_o(t_k)
    \times
    \left[
        \boldsymbol{x}_i(t_k)-\boldsymbol{p}_o(t_k)
    \right],
    \label{eq:rigid_vertex_velocity}
\end{equation}
where $\boldsymbol{v}_o$ and $\boldsymbol{\omega}_o$ denote the translational and
angular velocities of the object, respectively, $\boldsymbol{p}_o$ is its
world-space reference position, and $\boldsymbol{x}_i$ is the world-space vertex
position. The angular velocity is estimated from temporally adjacent rigid
transformations. This representation naturally captures local velocity
variations caused by object rotation, such as the motion of vehicle wheels.

For articulated or deformable objects, the complete Blender animation stack is first evaluated, and the vertex velocity ${v}_i(t_k)$ at an interior timestamp $t_k$ is estimated using a centered finite-difference approximation
\begin{equation}
    \boldsymbol{v}_i(t_k)
    =
    \frac{
        \boldsymbol{x}_i(t_{k+1})
        -
        \boldsymbol{x}_i(t_{k-1})
    }{
        t_{k+1}-t_{k-1}
    }.
    \label{eq:deformable_vertex_velocity}
\end{equation}
Since it is applied after rigid transformations,
skeletal articulation, and mesh deformation have been evaluated by Blender,
the resulting velocity implicitly contains both global motion and local
deformation. One-sided differences are used at the temporal boundaries.

For a ray intersecting a triangle formed by vertices $i_1$, $i_2$, and
$i_3$, the local surface velocity is obtained by barycentric interpolation
\cite{moller_trumbore_1997},
\begin{equation}
  \boldsymbol{v}_h = \sum_{q=1}^{3} \beta_q \boldsymbol{v}_{i_q},
  \quad \text{s.t.} \quad \sum_{q=1}^{3} \beta_q = 1,\ \beta_q \ge 0,
  \label{eq:bary_vel}
\end{equation}
where $\beta_q$ denotes the barycentric coordinate associated with the
$q$-th triangle vertex. 

To make this motion information directly accessible during propagation,
Since Sionna RT is built on Mitsuba 3 \cite{mitsuba3}, AuRaSim extends the corresponding Mitsuba mesh representation with a custom per-vertex velocity attribute. The world-space velocity field is transferred together with the updated mesh geometry, allowing the Path Solver to retrieve
and interpolate the local velocity directly at each ray--surface interaction.
The resulting hit-point velocity $\boldsymbol{v}_h^{(\ell)}$ of $\ell$-th interaction point along a propagation path is subsequently used in the motion-aware Doppler formulation described in Sec.~\ref{sec:path_doppler}.
\subsection{Path Solver}
\label{sec:path_tracer}

The AuRaSim Path Solver pipeline is illustrated in Fig.~\ref{fig:path_tracer}. At each timestamp, the current dynamic scene state is passed to the Path Solver, which combines \gls{sbr} with image-method refinement to determine valid propagation paths and subsequently evaluates their complex electromagnetic fields. AuRaSim adapts the Sionna RT pipeline to automotive radar simulation by introducing radar-oriented ray sampling and filtering, hit-point-level motion information, and an explicit path representation for waveform-level signal synthesis.


\subsubsection{FOV Ray Sampling}
\label{sec:fov_ray_sampling}

Building on the spherical Fibonacci sampling described in Sec.~\ref{sec:ray_sampling}, AuRaSim restricts the initial ray distribution to the radar-relevant \gls{fov}. Instead of generating rays over the
complete sphere and subsequently discarding those outside the radar field of
view, AuRaSim directly maps the Fibonacci sequence into the admissible region
$\Omega_{\mathrm{FOV}}$. Let
$\theta\in[\theta_{\min},\theta_{\max}]$ and
$\phi\in[\phi_{\min},\phi_{\max}]$ denote the configured polar and azimuth
ranges. The modified angles are

\[
    \theta_n'
    =
    \arccos\left[
        \cos\theta_\mathrm{\min}
        -
        \frac{n}{N_{\mathrm{ray}}-1}
        \left(
            \cos\theta_\mathrm{\min}
            -
            \cos\theta_\mathrm{\max}
        \right)
    \right],
    \label{eq:fov_theta}
\]
and
\[
    \phi_n'
    =
    \phi_\mathrm{\min}
    +
    \left(
        \phi_\mathrm{\max}-\phi_\mathrm{\min}
    \right)
    \mathrm{frac}\left(
        \frac{n}{\phi_g}
    \right).
    \label{eq:fov_phi}
\]

The initial ray directions are then obtained from
\eqref{eq:sionna_fibonacci} by replacing
$\theta_n$ and $\phi_n$ with
$\theta_n'$ and
$\phi_n'$, respectively.
The nonlinear mapping of $\theta$ preserves approximately uniform
solid-angle sampling within the restricted angular domain.

\begin{figure}[t]
    \centering
    \includegraphics[width=0.96\columnwidth]{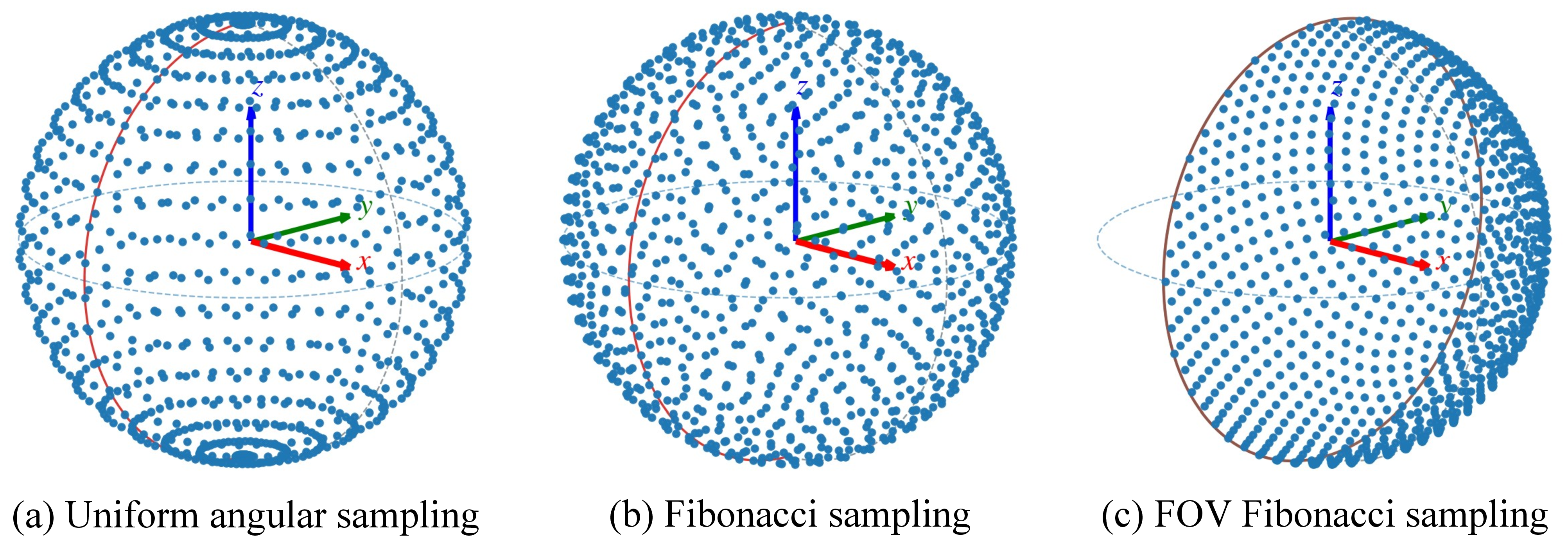}
    \caption{Illustration of ray sampling. Uniform angular sampling exhibits
    nonuniform point density over the sphere, while spherical Fibonacci sampling
    provides a more uniform full-sphere distribution. The proposed FOV-aware
    Fibonacci sampling concentrates the same number of rays directly within the
    radar field of view.}
    \label{fig:ray_sampling}
\end{figure}

Because AuRaSim changes the sampling domain itself rather than applying a
post-hoc angular filter, all $N_{\mathrm{ray}}$ samples remain available for
the radar-relevant region. For the same number of launched rays, this
increases the angular sampling density within the radar field of view, as
illustrated in Fig.~\ref{fig:ray_sampling}. Conversely, achieving the same
in-FOV sampling density with full-sphere sampling would require substantially
more rays and therefore higher ray-tracing cost.

As in Sec.~\ref{sec:ray_sampling}, each launched ray represents a finite solid angle. Restricting the sampling domain therefore requires updating this per-ray solid-angle weight. For the rectangular angular region considered above,
the corresponding
solid-angle weight per launched ray is

\begin{equation}
    \Delta\Omega_{\mathrm{FOV}}=
    \frac{
        |\left(
        \phi_\mathrm{\max}-\phi_\mathrm{\min}
    \right)
    \left(
        \cos\theta_\mathrm{\min}
        -
        \cos\theta_\mathrm{\max}
    \right)|
    }{
        N_{\mathrm{ray}}
    }.
\label{eq:fov_solid_angle}
\end{equation}
The propagation weights are therefore evaluated using
$\Delta\Omega_{\mathrm{FOV}}$, such that concentrating the ray samples
within the radar field of view improves sampling efficiency without
artificially changing the represented electromagnetic energy.

\subsubsection{Path Filtering}
\label{sec:path_filtering}

Before path refinement and electromagnetic field evaluation, AuRaSim applies
a radar-specific filtering stage to discard propagation paths that cannot
contribute to the radar observation. For each candidate path $p$, the arrival
direction and propagation distance are checked against the configured receive
field of view and maximum distance threshold. The path is retained only if
\[
    \mathbb{I}_{\mathrm{filter}}(p)
    =
    \mathbb{I}
    \left[
        -{\boldsymbol{k}}^{\mathrm{arr}}_p
        \in \Omega_{\mathrm{rx}}
    \right]
    \mathbb{I}
    \left[
        d_p \leq d_{\max}
    \right],
    \label{eq:path_filter}
\]
where ${\boldsymbol{k}}^{\mathrm{arr}}_p$ denotes the arrival direction of path $p$, $\Omega_{\mathrm{rx}}$ is the admissible receive angular region, $d_p$ is the propagation length, and $d_{\max}$ is the configured maximum length. Paths that violate either condition are discarded to avoid unnecessary path refinement and field evaluation
for \gls{sbr} candidates outside the useful radar observation region.

\subsubsection{Hit-Point Doppler}
\label{sec:path_doppler}

Sionna RT computes path Doppler by accumulating the velocity contributions of the objects intersected along the propagation path. In its standard mobility formulation, each scene object is associated with a single velocity vector $\boldsymbol{v}(o)$, which is applied identically to all interaction points on that object \cite{sionna}. This object-level approximation cannot represent the spatially varying motion of rotating, articulated, or deformable meshes.

AuRaSim replaces $\boldsymbol{v}(o)$ with the local hit-point velocity $\boldsymbol{v}_h^{(\ell)}$ at the $\ell$-th interaction point, described in Sec.~\ref{sec:vertex_velocity}. For a resolved propagation path $p$ containing $L$ interaction points and therefore $L+1$ propagation segments, the total first-order Doppler frequency $\nu_p$ is

\[
    \nu_p
    =
    \frac{1}{\lambda_0}
    \left[
        \boldsymbol{v}_{\mathrm{tx}}^{\mathsf T}
        {\boldsymbol{k}}_p^{(0)}
        -
        \boldsymbol{v}_{\mathrm{rx}}^{\mathsf T}
        {\boldsymbol{k}}_p^{(L)}
        +
        \sum_{\ell=1}^{L}
        \left(\boldsymbol{v}_h^{(\ell)}\right)^{\mathsf T}
        \left(
            {\boldsymbol{k}}_p^{(\ell)}
            -
            {\boldsymbol{k}}_p^{(\ell-1)}
        \right)
    \right].
\]
Here, $\boldsymbol{v}_{\mathrm{tx}}$ and $\boldsymbol{v}_{\mathrm{rx}}$ denote the transmitter and receiver velocities, respectively, while ${\boldsymbol{k}}_p^{(\ell)}$, $\ell=0,\ldots, L$, denotes the unit propagation direction of the corresponding path segment. Thus, the index $\ell=0$ denotes the segment leaving the transmitter, while $\ell=L$ denotes the final segment arriving at the
receiver. The first two terms account for transmitter and receiver motion, whereas each term in the summation represents the Doppler contribution of a moving interaction point through its velocity projected onto the change between the outgoing and incoming propagation directions. Replacing the object-level velocity with $\boldsymbol{v}_h^{(\ell)}$ therefore allows different locations on the same mesh to produce distinct Doppler shifts during waveform-level signal synthesis.

\subsubsection{Path Representation}
\label{sec:path_representation}

Each valid propagation path is exported as an explicit intermediate record
\[
    P_p
    =
    \left\{
        A_p,\,
        \tau_p,\,
        \nu_p,\,
        {\boldsymbol{k}}^{\mathrm{dep}}_p,\,
        {\boldsymbol{k}}^{\mathrm{arr}}_p,\,
        \mathbf{X}_p
    \right\},
\]
where $\tau_p$ denotes the propagation delay, ${\mathbf{k}}^{\mathrm{dep}}_p$ denotes the departure and arrival unit directions, and $\mathbf{X}_p$ contains the ordered interaction points along the path.

This path-level representation separates the computationally expensive propagation stage from waveform-dependent signal generation, allowing reuse for different waveform and acquisition configurations without repeating the ray-tracing computation and providing the input to the ADC Signal Synthesizer described in
Sec.~\ref{sec:adc_synthesizer}.

\subsection{ADC Signal Synthesizer}
\label{sec:adc_synthesizer}

The ADC Signal Synthesizer converts the resolved propagation paths into
complex sampled FMCW beat signals through coherent superposition. In contrast
to adding received powers independently, coherent synthesis preserves the
path phases and therefore captures constructive and destructive interference
between multipath components.

\subsubsection{MIMO Array Modeling}
\label{sec:mimo_array_modeling}

In explicit-array mode, propagation paths are evaluated for individual
transmit--receive pairs. In synthetic-array mode, ray tracing is performed
between the array reference positions, and the element-dependent phase
differences are subsequently reconstructed from the path departure and
arrival directions.

For path $p$, the phase compensation factor for transmitter $u$ and receiver
$r$ is
\begin{equation}
    c_{p,r,u}
    =
    \exp\left\{
        j\frac{2\pi}{\lambda_0}
        \left[
            \left(
                {\boldsymbol{k}}^{\mathrm{dep}}_p
            \right)^{\mathsf T}
            \boldsymbol{d}^{\mathrm{tx}}_u
            -
            \left(
                {\boldsymbol{k}}^{\mathrm{arr}}_p
            \right)^{\mathsf T}
            \boldsymbol{d}^{\mathrm{rx}}_r
        \right]
    \right\}.
    \label{eq:mimo_phase_compensation}
\end{equation}
The synthetic-array formulation assumes that the wavefront is approximately
planar across the array aperture, such that the propagation path
can be treated as common to all antenna elements and the element-dependent
path-length differences can be represented by phase shifts. This
approximation is valid when the array aperture is sufficiently small relative
to the propagation distance. When near-field effects or wavefront curvature
become significant, explicit transmit--receive tracing can instead be used
at increased computational cost.

\subsubsection{Coherent Multipath Synthesis}

Following the FMCW signal model introduced in
\eqref{eq:prelim_if_signal}, the resolved propagation paths are synthesized
coherently for each receive antenna. For receive antenna $r$, the sampled ADC signal $s_r[m,n]$ at chirp index $m$
and fast-time sample index $n$ is obtained by coherently summing all resolved propagation
paths,
\begin{equation}
\begin{split}
    s_r[m,n]
    =
    \sum_{p\in\mathcal{P}}
    &\underbrace{A_p}_{\text{path field}}
    \underbrace{c_{p,r,u(m)}}_{\text{MIMO}}
    \underbrace{e^{\jmath 2\pi f_c\tau_p}}_{\text{carrier phase}}
    \\
    &\cdot
    \underbrace{
        e^{\jmath 2\pi(\mu\tau_p-\nu_p)nT_s}
    }_{\text{fast time}}
    \underbrace{
        e^{-\jmath 2\pi\nu_p mT_r}
    }_{\text{slow time}} .
\end{split}
\label{eq:coherent_adc}
\end{equation}
Here, $T_s$ and $T_r$ denote the ADC sampling interval and chirp repetition
interval, respectively. For explicit-array simulation, the channel-specific path
coefficient $c_{p,r,u(m)}$ is not required.

By default, the path geometry is evaluated at a reference timestamp for each
radar frame and is assumed to remain approximately unchanged during the
corresponding coherent processing interval. Within the frame, dynamic objects
and sensors are assumed to follow locally constant velocities, while their
intra-frame motion is represented through the Doppler-induced phase
progression across chirps. However, the local velocities are updated between successive frames according to the animated scene, allowing time-varying micro-Doppler signatures from articulated objects to be simulated. 
This frame-level path evaluation substantially
reduces the computational cost while preserving coherent slow-time motion
effects. AuRaSim additionally supports chirp-level simulation, in which
the scene geometry and propagation paths are computed for individual
chirps, at the cost of increased ray-tracing time. 

\section{Experiments}
\label{sec:experiments}

\subsection{Data Preparation}
\label{sec:data_preparation}

\subsubsection{Dataset}
\label{sec:dataset}

The experiments are conducted on the dataset introduced by Gao
\emph{et al.} \cite{gao_rampcnn_2021}, which provides synchronized real radar measurements and RGB images for reconstructing and validating corresponding digital-twin scenarios. The full dataset contains approximately
19,800 synchronized raw radar frames and RGB images collected in campus,
road, and parking-lot environments. Three common road-user categories are
considered in this work: pedestrians, cyclists and cars.

\subsubsection{Digital-Twin Reconstruction}
\label{sec:Digital-Twin Reconstruction}
For signal-level evaluation, representative sequences from the dataset are reconstructed as corresponding digital-twin scenarios. The static environment is manually created and aligned with the radar coordinate system. Dynamic objects are detected from the synchronized RGB images using
the WildDet3D \cite{wilddet3d}, followed by temporal association, tracking, and trajectory smoothing. The corresponding 3D assets are
instantiated and animated according to the recovered trajectories, resulting
in temporally synchronized digital-twin scenes as described in
Sec.~\ref{sec:scenario_builder}. AuRaSim is then applied to these reconstructed scenes, and the simulated radar data are compared with the corresponding measured radar data. These scenes are subsequently
used for the real-to-sim validation and ablation studies presented in the
following sections. 

\subsubsection{Parameters}
\label{sec:parameters}
The measurements were acquired using a Texas Instruments AWR1843 radar
operating at 77~GHz. The radar configuration is summarized in
Table~\ref{tab:dataset_radar_config}. The same reconstructed scene geometry, antenna layout, antenna
pattern, waveform, and sampling configuration are reproduced
in AuRaSim and RadarSimPy \cite{radarsimpy} which is used as a physics-based baseline for comparison with AuRaSim. Both configurations are chosen to closely match the measurement setup. The ray-tracing backend of RadarSimPy controls spatial sampling
through a ray-density parameter rather than an explicit launched-ray count.
For the baseline experiments with RadarSimPy, the density is set to
$\rho_{\mathrm{ray}}=\mathrm{0.1}$ rays per wavelength squared, with a
maximum reflection order of 3 and diffuse scattering enabled. This value was selected as the highest stable setting for the reconstructed scenes used in our experiments, as higher ray densities consistently resulted in \gls{gpu} execution failures on our test system. Since the backend is provided as a precompiled implementation, the underlying cause could not be further isolated or modified.

\begin{table}[t]
\centering
\caption{Radar configuration used for both the real measurements and AuRaSim.}
\label{tab:dataset_radar_config}
\footnotesize
\renewcommand{\arraystretch}{1.15}
\setlength{\tabcolsep}{4pt}

\resizebox{\columnwidth}{!}{%
\begin{tabular}{@{}lc|lc@{}}
\toprule
\textbf{Parameter} & \textbf{Value} &
\textbf{Parameter} & \textbf{Value} \\
\midrule

Radar frame rate
& 30 Hz
& Carrier frequency $f_c$
& 77 GHz \\

Tx/Rx antennas
& $2\,\mathrm{TX}\times4\,\mathrm{RX}$
& MIMO scheme
& TDM \\

ADC samples/chirp
& 128
& Chirps per frame
& 510 \\

ADC sampling rate $f_s$
& 4 MSps
& Effective bandwidth $B$
& 0.67 GHz \\

Frequency slope $\mu$
& $21\,\mathrm{MHz}/\mu\mathrm{s}$
& Chirp duration
& $\approx60\,\mu\mathrm{s}$ \\

\bottomrule
\end{tabular}
}
\end{table}

For the main experiments, the path solver launches $10$ million initial rays
within a \gls{fov} of $\pm 90^\circ$ in both azimuth and elevation,
with a maximum interaction depth of 3. All simulations are performed on an NVIDIA RTX~3060 \gls{gpu}. 

\subsection{Evaluation Metrics}
\label{sec:evaluation_protocol}

For all signal-level comparisons, the measured and simulated \gls{adc} data are
processed using the same range--Doppler pipeline. The resulting RD power maps are independently normalized, converted to logarithmic scale, and clipped to a range of $-60$~dB to $0$~dB. The normalized dB values are subsequently mapped to $[0,1]$ for quantitative evaluation. Let
$\widetilde{M}^{\mathrm{real}}_{i,j}$ and
$\widetilde{M}^{\mathrm{sim}}_{i,j}$ denote the normalized log-power values of the measured and simulated \gls{rd} maps at range bin $i$ and Doppler bin $j$, respectively.
The root-mean-square error (RMSE) is
\begin{equation}
    \mathrm{RMSE}
    =
    \sqrt{
        \frac{1}{N_RN_D}
        \sum_{i=1}^{N_R}
        \sum_{j=1}^{N_D}
        \left(
            \widetilde{M}^{\mathrm{real}}_{i,j}
            -
            \widetilde{M}^{\mathrm{sim}}_{i,j}
        \right)^2
    },
    \label{eq:rd_rmse}
\end{equation}
and the \gls{psnr} is
\begin{equation}
    \mathrm{PSNR}
    =
    20\log_{10}
    \left(
        \frac{1}{\mathrm{RMSE}}
    \right).
    \label{eq:rd_psnr}
\end{equation}
Here, $N_R$ and $N_D$ denote the numbers of range and Doppler bins,
respectively. Lower RMSE and higher \gls{psnr} indicate closer agreement with the
measured radar response.
\subsection{Qualitative Results}
\label{sec:qualitative_results}

\begin{figure*}[t]
    \centering
    \includegraphics[width=\textwidth]{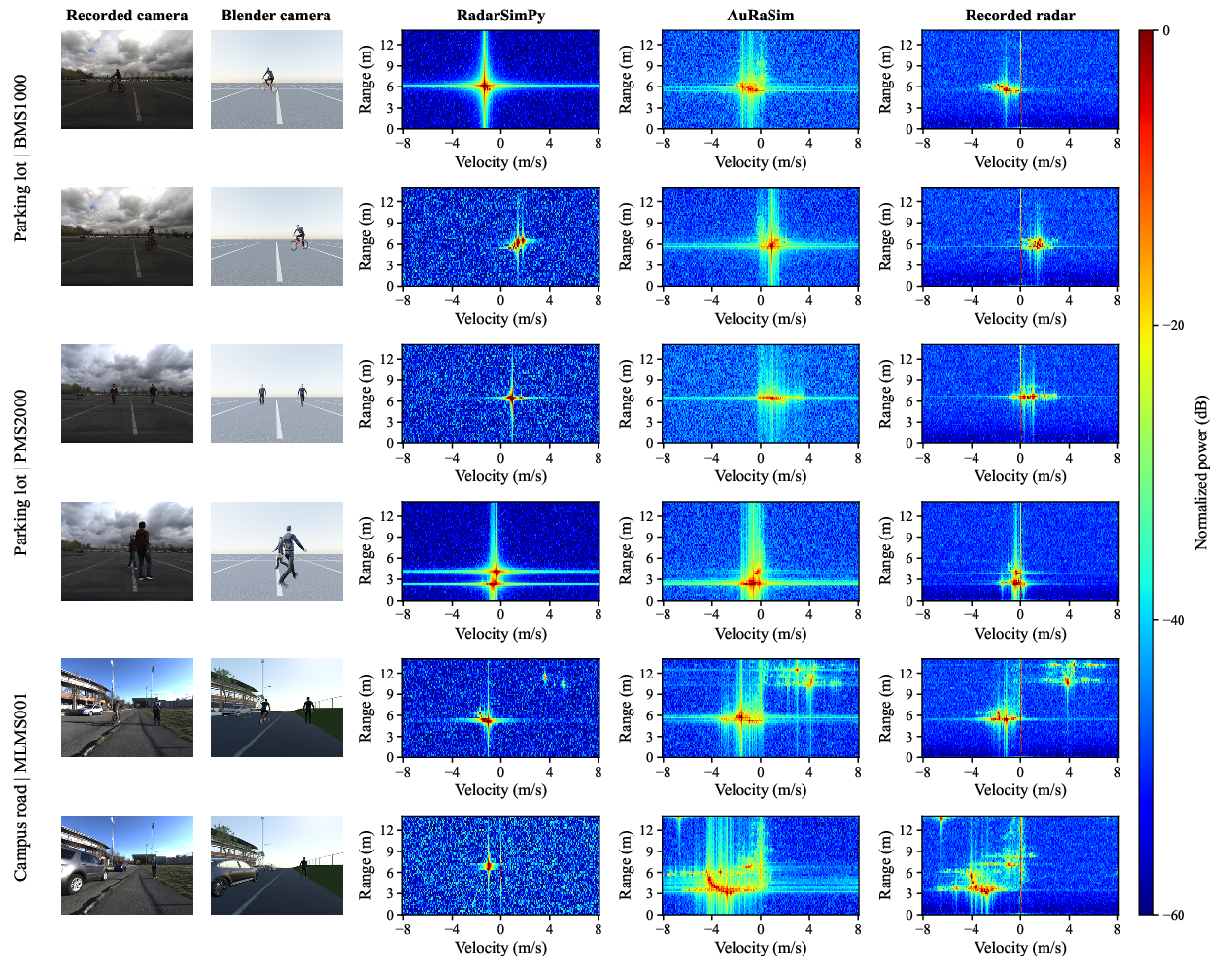}
    \caption{Representative qualitative comparison of real and simulated data. From left to right: real camera image, rendered digital-twin view, RadarSimPy RD map, AuRaSim RD map, and measured RD map.
    We show two representative frames for each sequence. For all RD maps, the horizontal axis corresponds
    to radial velocity, with negative velocities on the left (moving away) and
    positive velocities on the right (approaching), while the vertical axis corresponds
    to range, with range increasing from bottom to top. The color
    intensity represents the normalized radar power in dB.}
    \label{fig:rd_comparison}
\end{figure*}

Figure~\ref{fig:rd_comparison} presents representative frames from the three
evaluation sequences. The real camera images and the corresponding rendered
views illustrate the spatial alignment of the reconstructed digital twins,
while the remaining columns compare the RD maps generated by RadarSimPy and
AuRaSim with the measured radar data.

RadarSimPy generally recovers the approximate range and Doppler locations of
the dominant objects, demonstrating that the reconstructed scene geometry
provides a meaningful basis for radar simulation. However, several examples
show \gls{rd} magnitude distributions that differ from the measured \gls{rd} magnitude distributions. AuRaSim produces target responses that more closely resemble the
measured \gls{rd} structures, particularly in reproducing the range–Doppler extent and magnitude distributions of extended targets, including scenes with multiple independently moving objects. The
difference is most apparent in \texttt{MLMS001}, where more dynamic objects exist.

\begin{figure*}[t]
    \centering
    \includegraphics[width=0.985\textwidth]{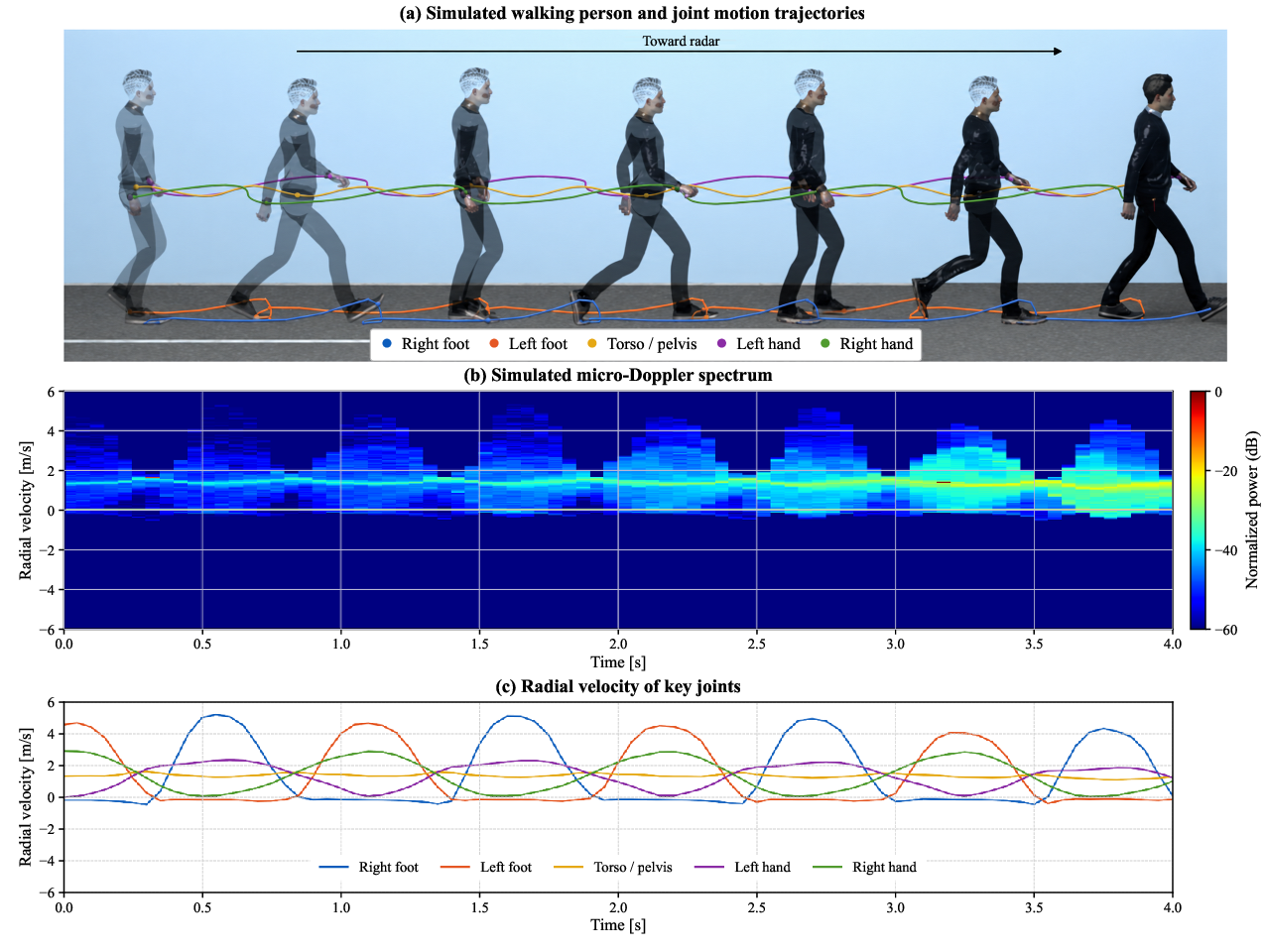}
    \caption{Micro-Doppler analysis of a simulated walking pedestrian using
    AuRaSim. (a) Reconstructed walking motion and trajectories of selected body
    joints while the pedestrian moves toward the radar. (b) Simulated
    micro-Doppler spectrum generated from the dynamic scene. (c) Radial
    velocities of representative body joints. The periodic Doppler spread is consistent with the articulated limb motion of the pedestrian.}
    \label{fig:micro_doppler}
\end{figure*}

Figure~\ref{fig:micro_doppler} demonstrates that AuRaSim can reproduce
micro-Doppler signatures arising from articulated human motion. The animated
pedestrian exhibits periodic limb motion and radial velocities, resulting in time-varying Doppler spreads around the main body
return. In particular, the larger velocity excursions of the limbs generate
the outer spectral components, while the torso remains concentrated near the
dominant body velocity. This result shows that AuRaSim preserves local
body-part motion in the radar signal rather than representing the pedestrian
as a single rigid target.

\subsection{Quantitative Results}
\label{sec:rd_results}

Table~\ref{tab:rd_quantitative} reports the quantitative agreement between
the measured RD maps and the results generated by RadarSimPy and AuRaSim. AuRaSim consistently achieves lower RMSE and higher \gls{psnr} in all three
sequences. The largest improvement is observed for \texttt{MLMS001}, where AuRaSim reduces the
RMSE from $0.233$ to $0.111$ and increases the \gls{psnr} from $12.90$ to
$19.40$~dB. Averaged over the three sequences, AuRaSim reduces RMSE by approximately $38.8\%$ and improves \gls{psnr} by $3.50$~dB compared with RadarSimPy.

\begin{table}[t]
\centering
\caption{Quantitative comparison between measured and simulated
range--Doppler maps in terms of RMSE and \gls{psnr}. $\downarrow$ denotes lower is
better and $\uparrow$ denotes higher is better.}
\label{tab:rd_quantitative}
\small
\renewcommand{\arraystretch}{1.12}
\begin{tabular}{llcc}
\toprule
\textbf{Scenario} &
\textbf{Method} &
\textbf{RMSE $\downarrow$} &
\textbf{PSNR [dB] $\uparrow$} \\
\midrule

\multirow{2}{*}{BMS1000}
& RadarSimPy~\cite{radarsimpy}
& 0.298
& 11.94 \\

& \textbf{AuRaSim}
& \textbf{0.172}
& \textbf{15.27} \\

\midrule

\multirow{2}{*}{PMS2000}
& RadarSimPy~\cite{radarsimpy}
& 0.189
& 15.46 \\

& \textbf{AuRaSim}
& \textbf{0.159}
& \textbf{16.11} \\

\midrule

\multirow{2}{*}{MLMS001}
& RadarSimPy~\cite{radarsimpy}
& 0.233
& 12.90 \\

& \textbf{AuRaSim}
& \textbf{0.111}
& \textbf{19.40} \\

\midrule

\multirow{2}{*}{Average}
& RadarSimPy~\cite{radarsimpy}
& 0.240
& 13.43 \\

& \textbf{AuRaSim}
& \textbf{0.147}
& \textbf{16.93} \\

\bottomrule
\end{tabular}
\end{table}

We attribute this improvement to two main differences. First, AuRaSim
models hit-point Doppler, allowing different parts of objects to
contribute distinct radial velocities, whereas the RadarSimPy baseline uses a
target-level motion representation. Second, AuRaSim explicitly controls the
number and angular distribution of launched rays and concentrates the samples
within the radar FOV, while RadarSimPy controls ray sampling through a density
parameter expressed in rays per wavelength squared. Although comparable
maximum interaction depths are used, the effective ray budgets,  internal path-search, and
field-evaluation procedures of the two solvers differ and therefore do not
permit an isolated comparison of numerical ray-tracing accuracy. The reported
results should thus be interpreted as an end-to-end comparison of the
resulting radar signals under matched reconstructed scenes and radar
configurations. The
influence of ray count and FOV-aware sampling is evaluated separately in
Sec.~\ref{sec:ablation}.

\subsection{Ablation Study}
\label{sec:ablation}

\begin{figure*}[t]
    \centering
    \includegraphics[width=0.99\textwidth]{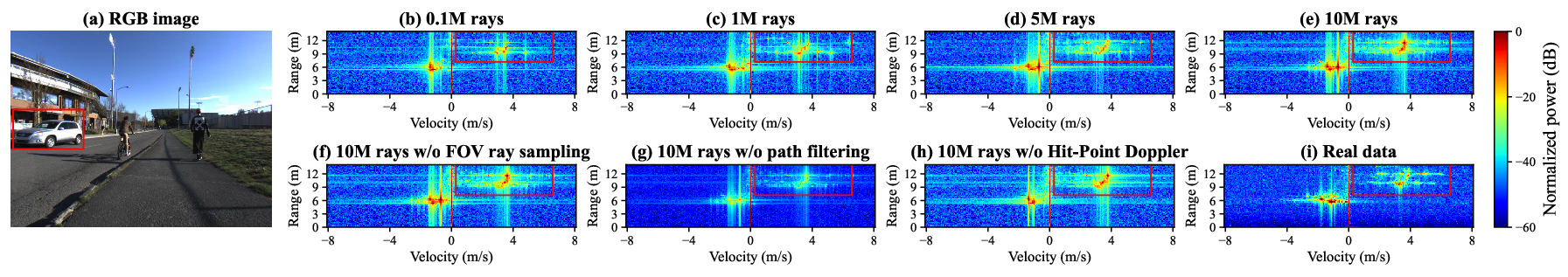}
    \caption{Qualitative ablation results on the \texttt{MLMS001} scenario.
    (a) RGB reference image. (b)--(e) Simulated range--Doppler maps using
    0.1, 1, 5, and 10 million rays, respectively. (f) Result with 10 million
    rays without the proposed FOV-aware ray sampling. (g) Result without path
    filtering. (h) Result using the conventional object-level Doppler model
    instead of the proposed local hit-point velocity formulation. (i) Measured
    radar data. The red boxes highlight the range--Doppler region associated
    with the vehicle-wheel response and are used to visualize differences in
    the corresponding micro-Doppler structure.}
    \label{fig:ablation_ray_count}
\end{figure*}

The ablation study investigates five key components of AuRaSim: the number of launched rays, FOV-aware ray sampling, path filtering, local hit-point Doppler modeling, and \gls{mimo} array modeling. The first four components are evaluated on the reconstructed \texttt{MLMS001} scenario using the same radar configuration and signal-processing pipeline. Table~\ref{tab:ablation} summarizes their quantitative results, while Fig.~\ref{fig:ablation_ray_count} presents the corresponding \gls{rd} maps. The MIMO array modeling is evaluated separately across all three sequences using \gls{ra} maps. Since the computational cost and reconstruction quality also depend on reconstruction quality and simulation settings, the reported results should be interpreted as comparisons within their respective experimental configurations rather than absolute simulator benchmarks.

\begin{table*}[t]
\centering
\caption{Ablation study on the \texttt{MLMS001} scenario. \gls{psnr} measures the
similarity between simulated and measured range--Doppler maps, while runtime
reports the corresponding ray-tracing time per radar frame.}
\label{tab:ablation}
\small
\renewcommand{\arraystretch}{1.12}
\begin{tabular}{ccccccc}
\toprule
\textbf{Exp.} &
\textbf{Rays [$\times 10^6$]} &
\textbf{\nameref{sec:fov_ray_sampling}} &
\textbf{\nameref{sec:path_filtering}} &
\textbf{\nameref{sec:path_doppler}} &
\textbf{PSNR [dB] $\uparrow$} &
\textbf{Runtime [s/frame] $\downarrow$} \\
\midrule

E1 & 0.1 & $\checkmark$ & $\checkmark$ & $\checkmark$ & 16.81 & 0.672 \\
E2 & 1   & $\checkmark$ & $\checkmark$ & $\checkmark$ & 18.21 & 0.690 \\
E3 & 5   & $\checkmark$ & $\checkmark$ & $\checkmark$ & 18.71 & 0.774 \\
E4 & 10  & $\checkmark$ & $\checkmark$ & $\checkmark$ & 19.40 & 0.838 \\
E5 & 10  & $\times$     & $\checkmark$ & $\checkmark$ & 18.59 & 0.859 \\
E6 & 10  & $\checkmark$ & $\times$     & $\checkmark$ & 15.46 & 0.826 \\
E7 & 10  & $\checkmark$ & $\checkmark$ & $\times$     & 17.52 & 0.837 \\

\bottomrule
\end{tabular}
\end{table*}

\subsubsection{Effect of the Number of Rays}

Experiments E1--E4 isolate the influence of the number of launched rays while keeping the remaining components unchanged. Increasing the ray count from 0.1 to 10 million consistently improves the agreement with the measured
range--Doppler map, increasing the \gls{psnr} from 16.81~dB to 19.40~dB. At the same time, the runtime increases from 0.672~s/frame to 0.838~s/frame.

The qualitative results in Fig.~\ref{fig:ablation_ray_count}(b)--(e) show the same tendency. At low sampling densities, only the dominant propagation
contributions are recovered, and parts of the target response remain weak or
incomplete. Increasing the number of rays resolves additional contributions
and progressively recovers the target structure. Once the dominant geometric
interactions of this scene are sufficiently sampled, however, the additional improvement becomes smaller while the ray-tracing cost continues to
increase. The results indicate that the 5--10 million range
already provides a dense representation for this particular scene.

\subsubsection{Effect of FOV Ray Sampling}

Experiments E4 and E5 provide the direct ablation of the proposed FOV-aware
sampling strategy at an identical budget of 10 million launched rays. Mapping the samples directly into the radar-relevant angular region increases the \gls{psnr} from 18.59 to 19.40~dB and slightly reduces the runtime from 0.859 to
0.838~s/frame. The difference arises because full-sphere sampling allocates a substantial fraction of the rays to directions outside the configured radar \gls{fov}.

For the full-sphere sampling, approximately half of the generated 
rays fall into the configured \gls{fov} used in this experiment. Consequently, E5 with 10 million full-sphere rays can also be compared with E3 using 5 million FOV-restricted rays as an approximate
effective-sampling-density comparison. Despite using only half as many
launched rays, E3 achieves a similar 18.71~dB compared with
18.59~dB for E5, while reducing the runtime from 0.859 to 0.774~s/frame,
corresponding to a reduction of approximately 0.085~s, or 9.9\%. This result
demonstrates that concentrating the ray budget within the relevant angular
region improves the efficiency of propagation sampling without increasing the total ray count.

\subsubsection{Effect of Path Filtering}

Experiments E4 and E6 isolate the effect of the path-filtering stage. Removing the filter reduces the \gls{psnr} substantially from 19.40 to 15.46~dB, while the runtime decreases slightly from 0.838 to 0.826~s/frame. This indicates that path filtering improves the simulation quality with a small computational cost. Otherwise, additional responses from paths outside the configured radar observation region, particularly around zero Doppler, reduce the relative contrast of the target responses.

\subsubsection{Effect of Doppler Modeling}

Finally, Experiments E4 and E7 evaluate the proposed local velocity-field
formulation while keeping the ray count, FOV sampling, and path filtering
unchanged. E7 uses the conventional object-level velocity representation,
whereas E4 evaluates the Doppler contribution using the local hit-point
velocity described in Sec.~\ref{sec:path_doppler}. The proposed model
improves the  from 17.52 to 19.40~dB, while the measured runtime remains
essentially unchanged at 0.837 and 0.838~s/frame, respectively.

The difference is particularly visible in the wheel-related region highlighted
by the red boxes in Fig.~\ref{fig:ablation_ray_count}. With the conventional
object-level Doppler formulation in Fig.~\ref{fig:ablation_ray_count}(h), the
vehicle body remains visible because its translational motion is still
represented. However, the additional Doppler spread associated with the
rotating wheel is substantially weaker because all interaction points on the
vehicle share the same object-level velocity. In contrast, the proposed
hit-point formulation assigns spatially varying velocities to individual surface interaction points, resulting in a more realistic wheel-induced micro-Doppler distribution that better agrees with the measured radar response in
Fig.~\ref{fig:ablation_ray_count}(i).

\subsubsection{Effect of MIMO Array Modeling}

\begin{figure*}[t]
    \centering
    \includegraphics[width=0.98\textwidth]{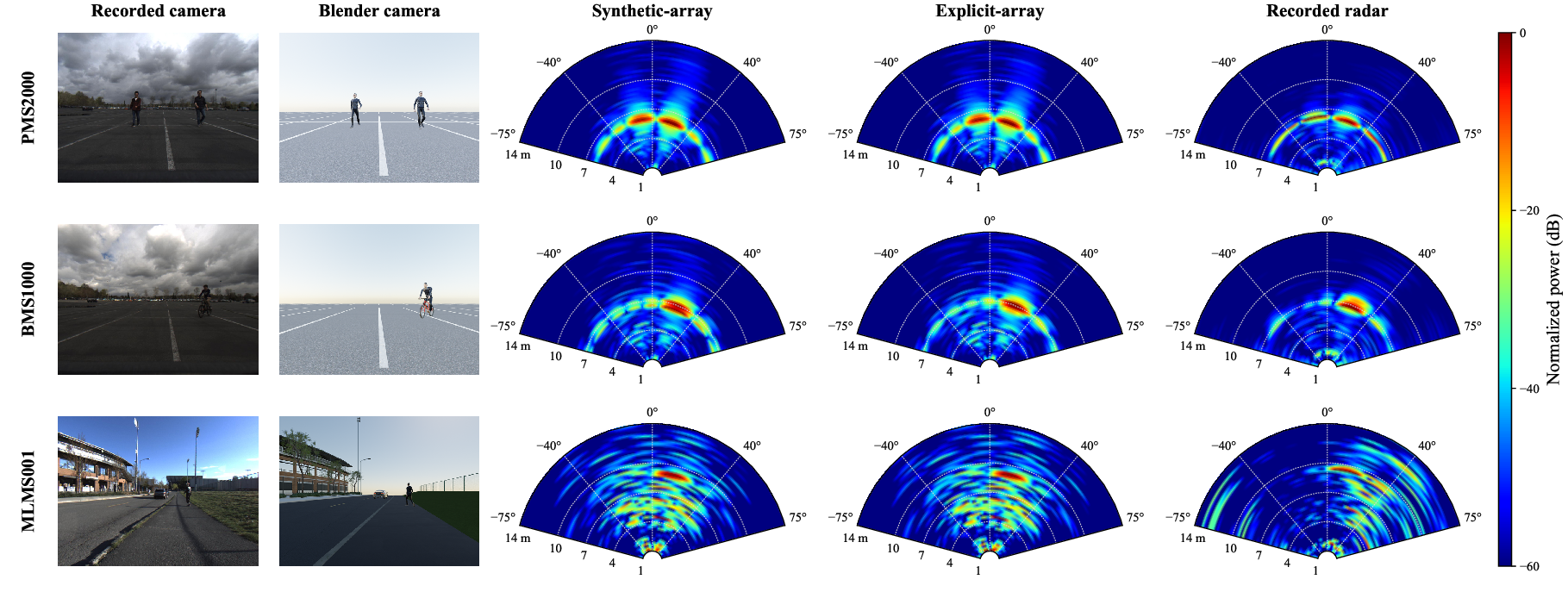}
    \caption{Qualitative comparison of synthetic-array and explicit-array MIMO modeling against measured range--azimuth (RA) maps. From top to bottom: \texttt{PMS2000}, \texttt{BMS1000}, and \texttt{MLMS001}. From left to right: recorded camera image, Blender-rendered scene, synthetic-array result, explicit-array result, and measured RA map. The radial and angular coordinates indicate range and azimuth, respectively. Each RA map is independently normalized and displayed on a relative power scale from $-60$ to $0$~dB.}
    \label{fig:ablation_mimo}
\end{figure*}

To evaluate the accuracy and computational efficiency of \gls{mimo} signal synthesis, we compare the synthetic-array and explicit-array modes described in Sec.~\ref{sec:mimo_array_modeling} against measured \gls{ra} maps from all three sequences using the corresponding \gls{psnr} and average ray-tracing time per frame. As shown in Table~\ref{tab:ablation_mimo} and Fig.~\ref{fig:ablation_mimo}, both modes produce similar angular responses, with \gls{psnr} differences below 0.2~dB. The phase-compensated synthetic-array mode therefore achieves comparable signal fidelity without sacrificing \gls{ra}-map reconstruction quality, while saving ray-tracing time by approximately 58\%. However, these results are obtained with the compact antenna array and target distances considered in this work, where the far-field approximation remains applicable. For larger array apertures or nearby targets, wavefront curvature and channel-dependent propagation delays may become significant. Evaluating these effects and the corresponding limitations remains future work.

\begin{table}[h]
\centering
\caption{Comparison of \gls{psnr} and runtime of MIMO array modeling modes with range--azimuth maps.}
\label{tab:ablation_mimo}
\footnotesize
\renewcommand{\arraystretch}{1.12}
\setlength{\tabcolsep}{4pt}
\begin{tabular}{llcc}
\toprule
\textbf{Scenario} &
\textbf{MIMO Mode} &
\textbf{PSNR [dB] $\uparrow$} &
\textbf{Runtime [s/frame] $\downarrow$} \\
\midrule

\multirow{2}{*}{BMS1000}
& Synthetic & 16.43  & \textbf{0.52} \\
& Explicit  & \textbf{16.46} & 1.26 \\

\midrule

\multirow{2}{*}{PMS2000}
& Synthetic & 17.73 & \textbf{0.52} \\
& Explicit  & \textbf{17.76} & 1.28 \\

\midrule

\multirow{2}{*}{MLMS001}
& Synthetic & 18.51 & \textbf{0.83} \\
& Explicit  & \textbf{18.70} & 2.07 \\

\midrule

\multirow{2}{*}{Average}
& Synthetic & 17.56 & \textbf{0.62} \\
& Explicit  & \textbf{17.64} & 1.54 \\

\bottomrule
\end{tabular}
\end{table}

\section{Conclusion}
\label{sec:conclusion}

This paper presented AuRaSim, an end-to-end automotive FMCW radar simulation
framework that connects dynamic digital-twin scenes, configurable MIMO radar
sensors, physics-based propagation, and waveform-level signal synthesis within
a common and traceable pipeline. Building on Blender and Sionna RT, AuRaSim supports static, rigid, and deformable scene geometry, configurable antenna patterns and radio materials, path-resolved electromagnetic propagation, motion-aware Doppler modeling and coherent multi-channel ADC generation. We
evaluated the simulator using digital twins reconstructed from real 77-GHz
automotive radar measurement scenarios. Quantitative comparison between
simulated and measured range--Doppler maps yielded an average RMSE of
$0.147$ and an average  of $16.93$~dB across the evaluated sequences.
These results indicate that AuRaSim can reproduce the dominant range–Doppler response structures observed in real measurements while retaining a
physically interpretable scene-to-signal mapping.

However, the fidelity of physics-based simulation remains affected by the
accuracy of electromagnetic material parameters. Sensor-specific effects such
as noise, interference, phase noise, calibration errors, and
adverse-weather phenomena are not explicitly modeled in the current
framework. In addition, high-order multipath and dense dynamic scenes remain
computationally demanding despite the proposed optimizations.

Future work will focus on measurement-driven calibration of material and
sensor parameters, more complete modeling of environmental effects, and
further acceleration of ray tracing and ADC synthesis. AuRaSim also provides
a basis for differentiable inverse radar rendering, physics-guided Sim2Real
transfer, and large-scale synthetic-data generation for downstream automotive
radar perception.

\bibliographystyle{IEEEtran}
\bibliography{reference}
\begin{IEEEbiography}[{\includegraphics[width=1in,height=1.25in,clip,keepaspectratio]{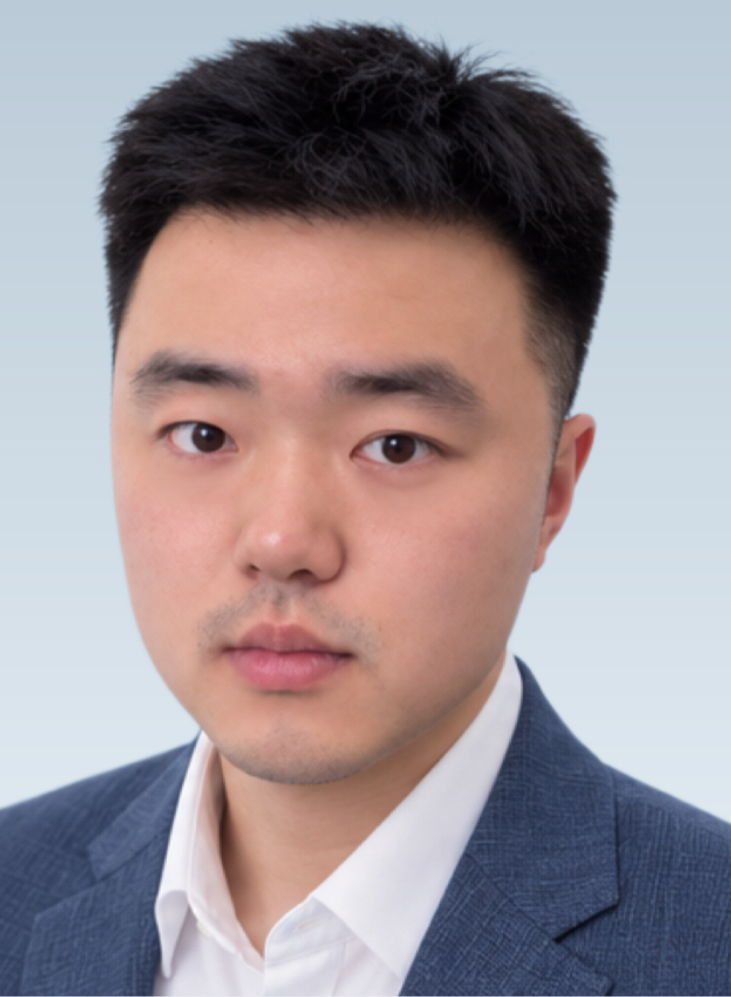}}]{Zhaoze Wang}
(Graduate Student Member, IEEE) received the B.S. degree in Mechanical Engineering from Harbin Engineering University, Harbin, China, in 2020, and the M.Sc. degree in Mechatronics from Friedrich-Alexander-Universität Erlangen-Nürnberg (FAU), Erlangen, Germany, in 2024. He is currently pursuing the Dr.-Ing. degree in Electrical Engineering with the Chair of Electronic Systems and Sensors, Brandenburg University of Technology, Cottbus-Senftenberg, Germany, and HELLA GmbH \& Company KGaA, Lippstadt, Germany. His research interests include physics-based automotive radar simulation, generative artificial intelligence, world models, and scene understanding for realistic and scalable radar data generation.\end{IEEEbiography}

\begin{IEEEbiography}[{\includegraphics[width=1in,height=1.25in,clip,keepaspectratio]{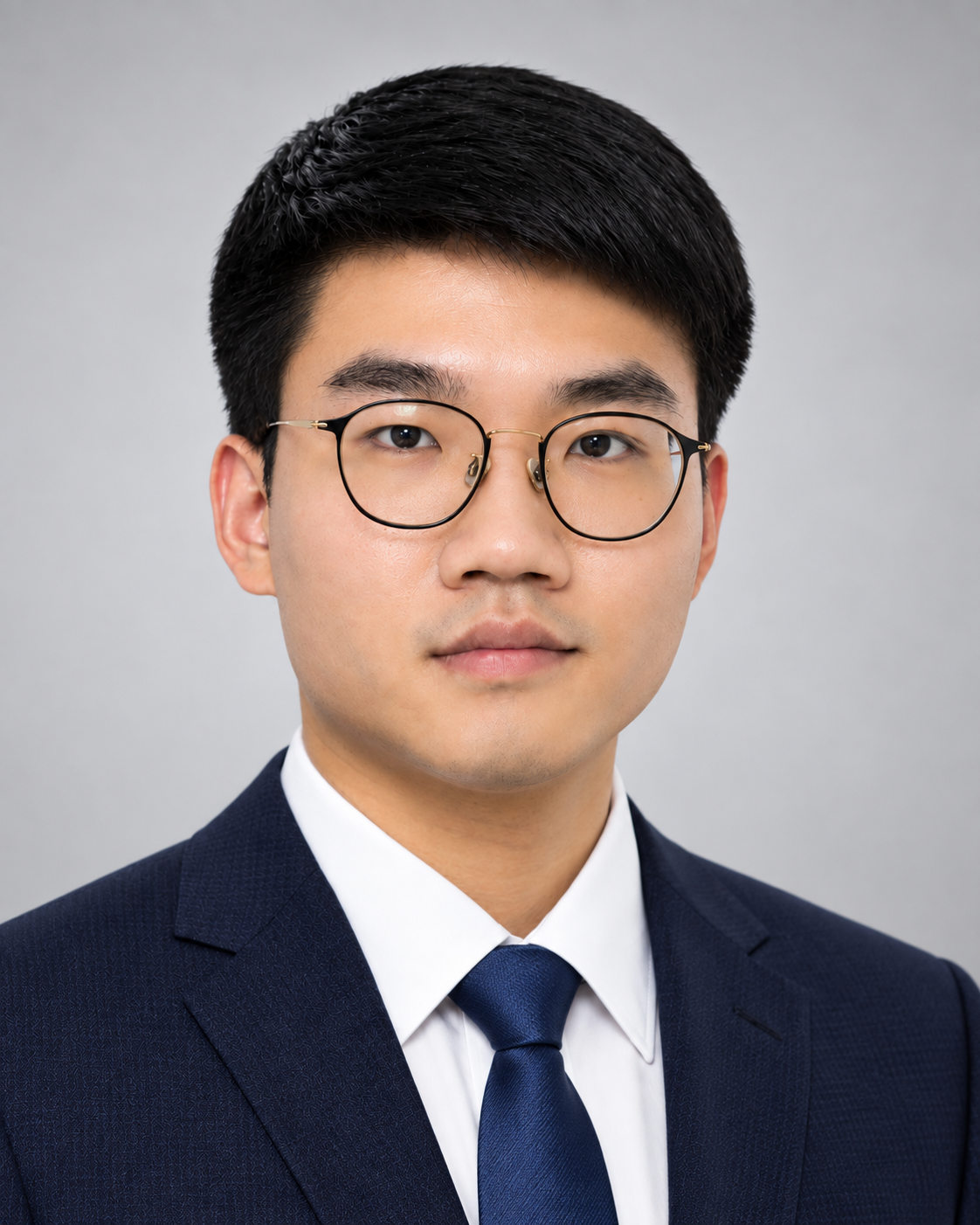}}]{Changxu Zhang}
(Graduate Student Member, IEEE) received the B.Eng. degree in automotive
engineering from Chang'an University, Xi'an, China, in 2020, and the M.Sc.
degree in automotive systems from Technische Universität Berlin, Berlin,
Germany, in 2024. He is currently pursuing the Ph.D. degree in electrical engineering while working with HELLA GmbH \& Co. KGaA, Lippstadt, Germany, in cooperation with Brandenburg University of Technology
Cottbus--Senftenberg, Cottbus, Germany. His research interests include deep
learning for automotive radar perception, sensor fusion, self-supervised
learning, and radar signal processing.
\end{IEEEbiography}

\begin{IEEEbiography}[{\includegraphics[width=1in,height=1.25in,clip,keepaspectratio]{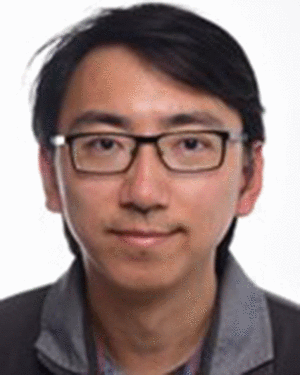}}]{Tai Fei}
(Senior Member, IEEE) received the B.Eng. degree from Shanghai Maritime
University, Shanghai, China, in 2005, and the Dipl.-Ing. and Dr.-Ing. degrees
in electrical engineering from TU Darmstadt, Darmstadt, Germany, in 2009 and
2014, respectively. He is an interim professor of computer vision and robotics
with the Dortmund University of Applied Sciences and Arts, Dortmund, Germany.
From 2014 to 2023, he was a development engineer with HELLA GmbH \& Co. KGaA,
where he worked on automotive radar. His research interests include automotive
radar, computer vision, robotics, and machine learning. He has authored more
than 90 publications and holds 11 granted patents.
\end{IEEEbiography}
 
\begin{IEEEbiography}[{\includegraphics[width=1in,height=1.25in,clip,keepaspectratio]{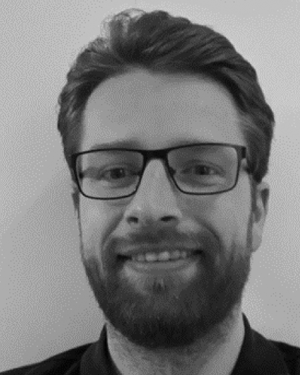}}]{Christopher Grimm}
(Graduate Student Member, IEEE) received the B.Eng. degree in automotive
engineering from the Dortmund University of Applied Sciences and Arts,
Dortmund, Germany, in 2013, and the M.Sc. degree in mechanical engineering
from the Technical University of Braunschweig, Braunschweig, Germany, in 2015.
He has been pursuing the Dr.-Ing. degree in electrical engineering with the
Department of Communications Engineering, University of Paderborn, Paderborn,
Germany, and HELLA GmbH \& Co. KGaA, Lippstadt, Germany, since 2016. Since
2020, he has been a development engineer with HELLA, where he develops
artificial-intelligence-based radar signal-processing algorithms. His research
interests include automotive radar signal processing, machine learning, and
computer vision.
\end{IEEEbiography}
 
\begin{IEEEbiography}[{\includegraphics[width=1in,height=1.25in,clip,keepaspectratio]{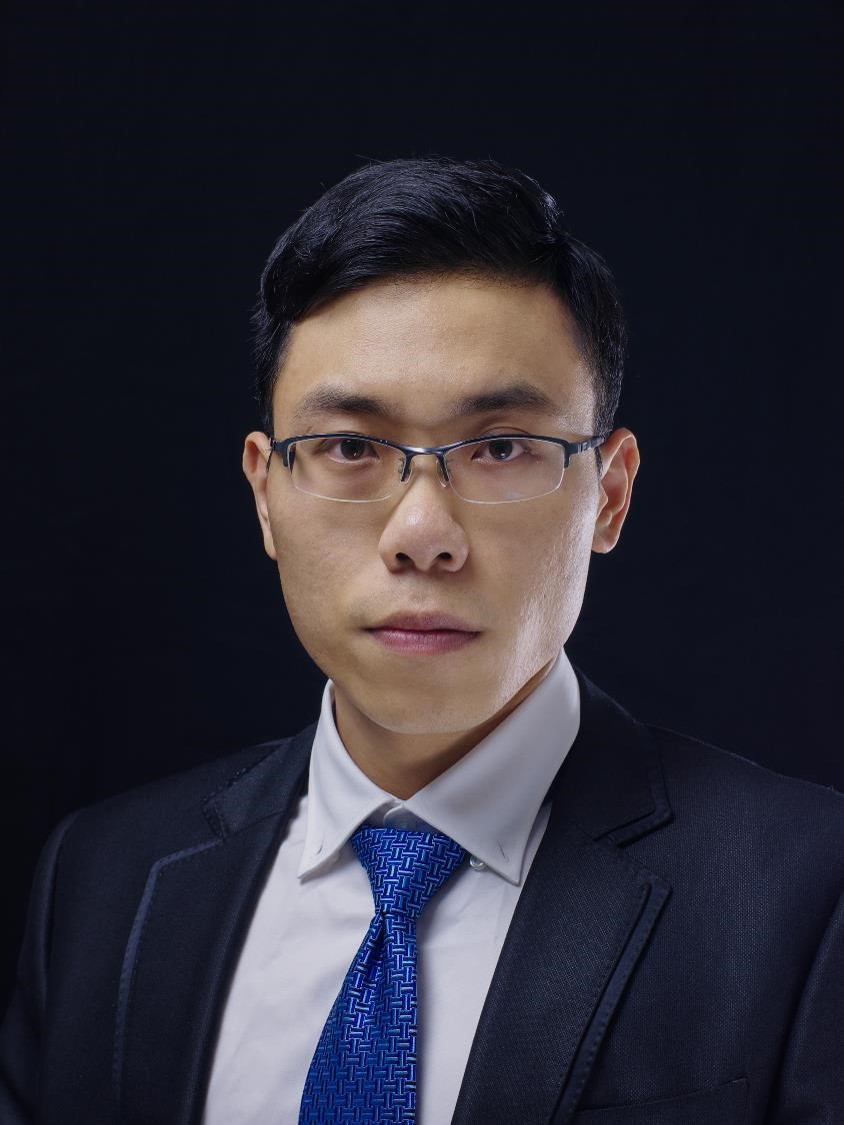}}]{Yi Jin}
(IEEE Member) received the M.Sc. degree in electromobility from the University of Stuttgart, Stuttgart, Germany, in 2019, and the Dr.-Ing. degrees
in electrical engineering from Friedrich-Alexander-Universität Erlangen-Nürnberg, Erlangen, Germany, in 2023. He is currently a radar perception engineer with Forvia Hella, Lippstadt, Germany. His research interests include automotive radar signal processing, radar perception and sensor fusion.
\end{IEEEbiography}
 
\begin{IEEEbiography}[{\includegraphics[width=1in,height=1.25in,clip,keepaspectratio]{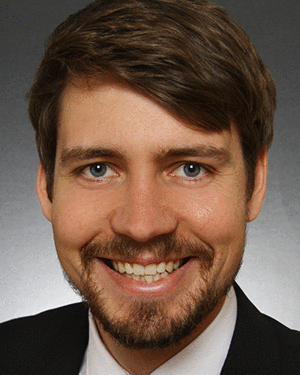}}]{Claas Tebruegge}
(Member, IEEE) received the B.Sc. degree in general engineering science and
the M.Sc. degree in electrical engineering from the Hamburg University of
Technology, Hamburg, Germany, in 2014 and 2016, respectively, and the Ph.D.
degree (Dr.~rer.~nat.) from Paderborn University, Paderborn, Germany, in 2022,
for research on vehicular visible-light communication. He joined HELLA GmbH
\& Co. KGaA in 2016 and has headed the E-LAB, the Research Institute for
Automotive Electronics, since 2022. His research interests include automotive
electronics, communication engineering, radar, signal processing,
electrification, and sustainable mobility.
\end{IEEEbiography}
 
\begin{IEEEbiography}[{\includegraphics[width=1in,height=1.25in,clip,keepaspectratio]{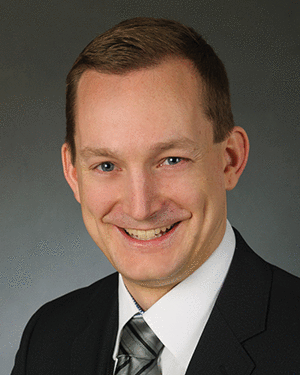}}]{Markus Gardill}
(Senior Member, IEEE) is a professor and head of the Chair of Electronic
Systems and Sensors at Brandenburg University of Technology
Cottbus--Senftenberg, Cottbus, Germany. He previously held a professorship in
satellite communication systems at the Julius-Maximilian University of
Würzburg, Würzburg, Germany, and held industry positions with Robert Bosch
GmbH, Bamberg, Germany, and InnoSenT GmbH, Donnersdorf, Germany. His research
interests include radar and communication systems, antenna-array design, and
signal-processing algorithms, with an emphasis on integrating signal
processing with microwave and electromagnetic engineering. He received the
MTT-S Outstanding Young Engineer Award in 2024.
\end{IEEEbiography}
 
\vfill

\end{document}